\documentclass[nohyperref]{article}

\usepackage[T1]{fontenc}
\usepackage{microtype}
\usepackage{graphicx}
\usepackage{subcaption}
\usepackage{booktabs} 
\usepackage{adjustbox}

\usepackage{hyperref}

\usepackage[accepted]{icml2023}

\usepackage{amsmath}
\usepackage{amssymb}
\usepackage{mathtools}
\usepackage{amsthm}

\usepackage[capitalize,noabbrev,nameinlink]{cleveref}

\theoremstyle{plain}

\theoremstyle{definition}

\theoremstyle{remark}

\usepackage[textsize=tiny]{todonotes}

\icmltitlerunning{Anti-Exploration by Random Network Distillation}

\begin{document}

\twocolumn[

\icmltitle{Anti-Exploration by Random Network Distillation}

\icmlsetsymbol{equal}{*}

\begin{icmlauthorlist}
\icmlauthor{Alexander Nikulin}{comp}
\icmlauthor{Vladislav Kurenkov}{comp}
\icmlauthor{Denis Tarasov}{comp}
\icmlauthor{Sergey Kolesnikov}{comp}
\end{icmlauthorlist}

\icmlaffiliation{comp}{Tinkoff, Moscow, Russia}

\icmlcorrespondingauthor{Alexander Nikulin}{a.p.nikulin@tinkoff.ai}

\icmlkeywords{ICML, Machine Learning, Reinforcement Learning, Offline Reinforcement Learning, Random Network Distillation, Epistemic Uncertainty}

\vskip 0.3in
]


\printAffiliationsAndNotice{} 
\begin{abstract}
Despite the success of Random Network Distillation (RND) in various domains, it was shown as not discriminative enough to be used as an uncertainty estimator for penalizing out-of-distribution actions in offline reinforcement learning. In this paper, we revisit these results and show that, with a naive choice of conditioning for the RND prior, it becomes infeasible for the actor to effectively minimize the anti-exploration bonus and discriminativity is not an issue. We show that this limitation can be avoided with conditioning based on Feature-wise Linear Modulation (FiLM), resulting in a simple and efficient ensemble-free algorithm based on Soft Actor-Critic. We evaluate it on the D4RL benchmark, showing that it is capable of achieving performance comparable to ensemble-based methods and outperforming ensemble-free approaches by a wide margin.    
\end{abstract}

\section{Introduction}
\label{intro}

\begin{figure}[h]
    \vskip 0.2in
    \centerline{\includegraphics[width=0.8\columnwidth]{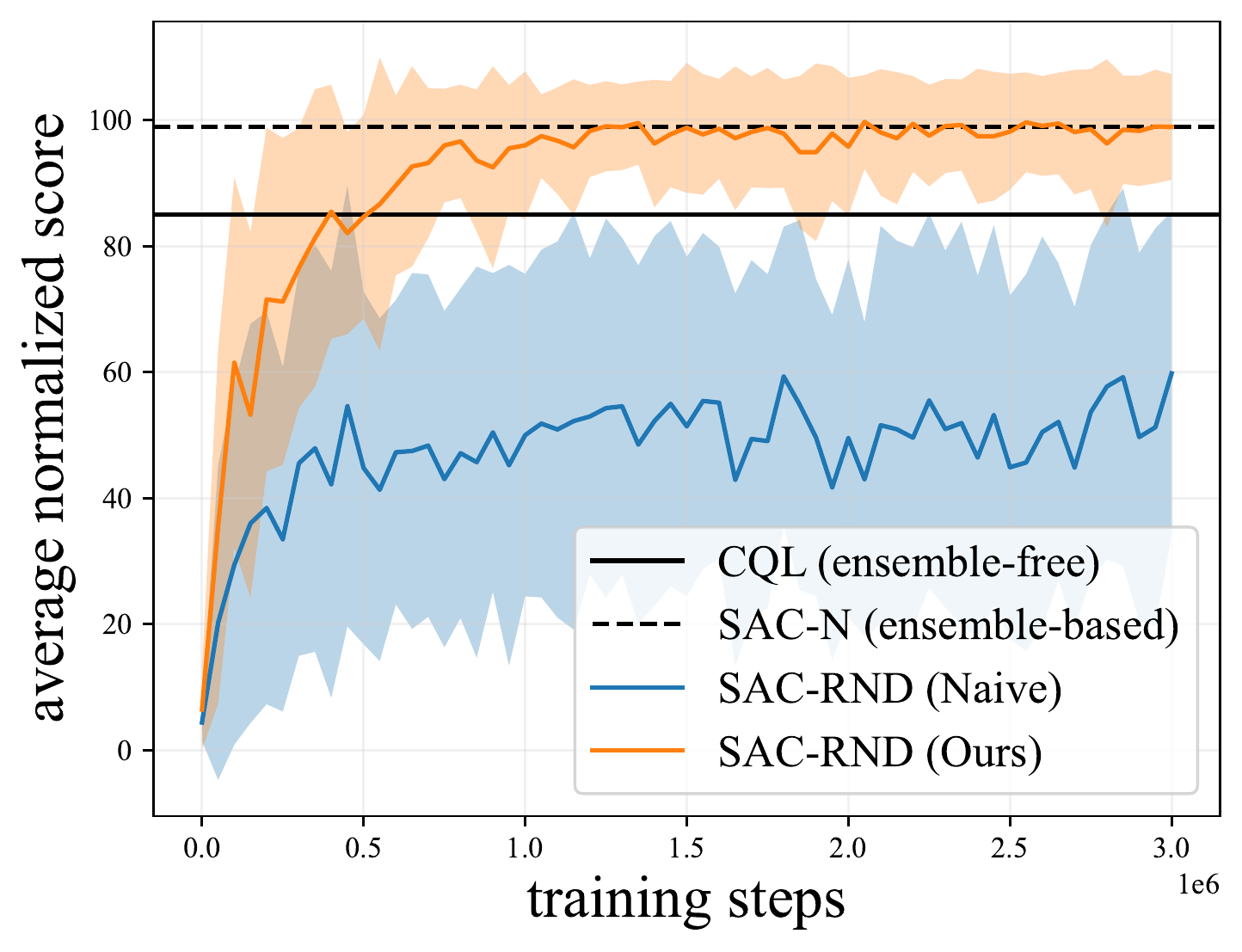}}
    \caption{Mean performance of SAC-RND variants on walker and hopper medium-* datasets, each averaged over 3 seeds. We plot performance for the naive version, which uses concatenation conditioning, and our final version, which is described in \cref{method}. We also plot the final scores for the ensemble-free CQL \citep{kumar2020conservative} and the ensemble-based SAC-N \citep{an2021uncertainty}. It can be seen that our version is a significant improvement over the naive version, achieving performance comparable to ensembles.}
    \label{fig:rnd-naive}
    \vskip -0.2in
\end{figure}

In recent years, significant success has been achieved in applying Reinforcement Learning (RL) to challenging and large-scale tasks such as Atari \citep{badia2020agent57}, Go \citep{schrittwieser2020mastering}, Dota 2 \citep{berner2019dota}, and Minecraft \citep{baker2022video}. However, the online nature of such RL algorithms makes it difficult to apply them in the real world, where online collection of large amounts of exploratory data may not be feasible for safety or financial reasons. Offline Reinforcement Learning \citep{levine2020offline} promises a more controllable and data-driven approach, focusing on algorithms that can learn from a fixed, pre-recorded dataset without requiring additional environment interactions.


The use of ensembles for uncertainty-based penalization has proven to be one of the most effective approaches for offline RL. Ensemble-based algorithms, such as SAC-N, EDAC \citep{an2021uncertainty}, and MSG \citep{ghasemipour2022so} currently achieve state-of-the-art results on most D4RL \cite{fu2020d4rl} datasets, outperforming ensemble-free methods by a wide margin. Unfortunately, in order to achieve the best performance, these algorithms may require tens or hundreds of ensemble members, leading to significant computational and memory overhead, as well as extended training duration \cite{nikulin2022q}.

Recent research \citep{yang2022rorl} has successfully reduced the ensemble size to tens of Q-networks in the worst-case scenarios. However, given the general trend for model scaling in offline RL \citep{kumar2022offline, reed2022generalist, lee2022multi}, efficiently training even ten Q-networks with 80 million parameters each is not feasible. Furthermore, \citet{ghasemipour2022so} showed that methods for efficient ensemble training found in supervised learning literature do not deliver performance comparable to naive ensembles and can even worsen the results. Thus, further research on efficient uncertainty estimation for offline RL is needed, with the goal of reducing the size of the ensemble as much as possible or even fully removing it.

In this work, we move away from ensembles and take an alternative approach to uncertainty estimation, proposing an efficient offline RL method with ensemble-free uncertainty estimation via Random Network Distillation (RND) \citep{burda2018exploration}. RND, a simple and fast ensemble competitor for epistemic uncertainty estimation \citep{ciosek2019conservative}, is an attractive choice for offline RL. However, previous research \citep{rezaeifar2022offline} found RND to be insufficiently discriminative for good results.

In our preliminary experiment (\cref{reproduction}), we show that RND is discriminative enough to detect OOD actions, which contradicts the previous study \citep{rezaeifar2022offline}. Nevertheless, our results show that the naive application of RND does indeed not lead to good results (see \cref{fig:rnd-naive}). Building upon these findings, we further simplify the problem and analyze the reasons for this issue (\cref{demonstration}). We discover that a naive choice of conditioning for the RND prior can hinder the minimization of the anti-exploration bonus by the actor, and that conditioning based on Feature-wise Linear Modulation (FiLM) \citep{perez2018film} is particularly effective in solving this problem. 

Based on our findings, we propose a new ensemble-free offline RL algorithm called \textbf{SAC-RND} (\cref{method}). We evaluate our method on the D4RL \citep{fu2020d4rl} benchmark (\cref{exp}), and show that SAC-RND achieves performance comparable to ensemble-based methods while outperforming ensemble-free approaches.

\section{Background}
\label{prelim}

\textbf{Offline Reinforcement Learning}. Reinforcement learning problem can be described as a Markov Decision Process (MDP) defined by the $\{\mathcal{S}, \mathcal{A}, \mathcal{P}, \mathcal{R}, \gamma\}$ tuple with state space $\mathcal{S} \subset \mathbb{R}^N$, action space $\mathcal{A} \subset \mathbb{R}^M$, transition dynamics $\mathcal{P} : \mathcal{S} \times \mathcal{A} \rightarrow \mathcal{S}$, reward function $\mathcal{R} : \mathcal{S} \times \mathcal{A} \rightarrow \mathbb{R}$, and a discount factor $\gamma$. The goal of reinforcement learning in an infinite horizon setting is to produce a policy $\pi(a | s)$ that maximizes the expected cumulative discounted return $\mathbb{E}_{\pi} [ \sum_{t=0}^{\infty} \gamma^t r(s_{t}, a_{t})]$. 

In offline reinforcement learning, a policy must be learned from a fixed dataset $\mathcal{D}$ collected under a different policy or mixture of policies, without any environment interaction. This setting poses unique fundamental challenges \cite{levine2020offline}, since the learning policy is unable to explore and has to deal with distributional shift and extrapolation errors \cite{fujimoto2019off} for actions not represented in the training dataset. 

\textbf{Offline RL as Anti-Exploration}. There are numerous approaches for offline RL, a substantial part of which constrain the learned policy to stay within the support of the training dataset, thus reducing \cite{kumar2020conservative} or avoiding \cite{kostrikov2021offline} extrapolation errors. For our work, it is essential to understand how such a constraint can be framed as \emph{anti-exploration} \cite{rezaeifar2022offline}. 

Similarly to online RL, where novelty bonuses are used as additive intrinsic rewards for effective exploration, in offline RL, novelty bonuses can induce conservatism, reducing the reward in unseen state-action pairs. Hence the name \emph{anti-exploration}, since the same approaches from exploration can be used, but a bonus is subtracted from the extrinsic reward instead of being added to it.

However, unlike online RL, subtracting a bonus from the raw reward would not be as useful, since the novelty bonus is, by design, close to zero for in-dataset state-action pairs. Therefore, it is more effective to apply it where the overestimation for OOD actions emerges — the temporal difference learning target:

\begin{equation}
\label{eq:q-target}
    r + \gamma \mathbb{E}_{a' \sim \pi(\cdot | s')} [Q(s', a') - \textcolor{blue}{b(s', a')}]
\end{equation}

where the actor is trained to maximize the expected Q-value, as is usually done in off-policy actor-critic algorithms \cite{lillicrap2015continuous, haarnoja2018soft}. It can be shown that, theoretically, these approaches are equivalent, but the latter is more suited for use in offline RL \cite{rezaeifar2022offline}. 

An illustrative example of how such framing can be effective are ensemble-based approaches such as SAC-N \& EDAC \cite{an2021uncertainty} and MSG \cite{ghasemipour2022so}, which currently outperform their ensemble-free counterparts by a large margin on most D4RL \cite{fu2020d4rl} benchmark datasets. For the anti-exploration bonus, these methods use ensemble disagreement as a proxy for epistemic uncertainty. However, a large number of ensemble members is usually required for a competitive result. 

\textbf{Random Network Distillation}. Random network distillation (RND) was first proposed in online RL \cite{burda2018exploration} as a simple and effective exploration bonus. 
To this day, RND is still considered a strong baseline for exploration that can work well even in stochastic environments, contrary to some more modern approaches \cite{jarrett2022curiosity}.

RND consists of two neural networks: a fixed and randomly initialized \emph{prior} network $\bar{f}_{\bar{\psi}}$, and a \emph{predictor} network $f_{\psi}$ which learns to predict the prior outputs on the training data:
\begin{equation}
\label{eq:online-rnd-loss}
 \lVert f_{\psi}(s) - \bar{f}_{\bar{\psi}}(s) \rVert_{2} ^ 2
\end{equation}

Both networks map states to embeddings in $\mathbb{R}^K$, and the gradient through \emph{prior} is disabled. The interpretation of the novelty is straightforward: with the sufficiently diverse \emph{prior}, the \emph{predictor} must learn to match embeddings on data points similar to the training dataset, while failing to predict on new examples. A bonus in such a case may simply be a prediction error, as in \cref{eq:online-rnd-loss}.

In a subsequent work, \citet{ciosek2019conservative} analyses the success of RND in a supervised setting, and shows that fitting random priors can be a competitive alternative to ensembles for estimating epistemic uncertainty. 



Note that in practice, the choice of predictor and prior having the same architecture and the estimation of novelty from states only are \emph{very common, but arbitrary}. Moreover, for offline RL, we are interested in estimating the novelty of an action conditioned on the state, which is why in our work RND depends on both: $f_{\psi}(s, a)$.


\textbf{Multiplicative Interactions}. The most common way to fuse two different streams of information is feature concatenation, which is straightforward but can be suboptimal \citep{dumoulin2018feature-wise}. \citet{jayakumar2020multiplicative} shows that multiplicative interactions provide a powerful inductive bias for fusing or conditioning from multiple streams and are superior in practice. We provide a brief review of those used in our work (excluding concatenation): gating, bilinear, and feature-wise linear modulation (FiLM). 


\textbf{Gating}. Simple conditioning with two linear layers and pointwise multiplication of the resulting features \citep{srivastava2019training}.
\begin{equation*}
f(a, s) = tanh(W_{1}a + b_{1}) \odot \sigma(W_{2}s + b_{2})
\end{equation*}

\textbf{Bilinear}. Bilinear layer in its most general form, as proposed by \citet{jayakumar2020multiplicative}.
\begin{equation*}
f(a, s) = s^{T}\mathbb{W}a + s^{T}\mathbb{U} + \mathbb{V}a + b
\end{equation*}
where $\mathbb{W}$ is a 3D tensor, $\mathbb{U}$, $\mathbb{V}$ are regular matrices and $b$ is a vector. However, in our work, we also use the implementation as in PyTorch \citep{paszke2019pytorch}, which does not learn $\mathbb{U}$, $\mathbb{V}$ by default.

\textbf{FiLM}. Special case of a bilinear layer 
with low-rank weight matrices \citep{perez2018film}.
\begin{equation*}
f(h, s) = \gamma(s) \odot h  + \beta(s)
\end{equation*}

Usually, FiLM operates on hidden activations $h$ before nonlinearity between layers. Thus, the main network takes $a$ as an input.

\section{Random Network Distillation is Discriminative Enough}
\label{reproduction}

To better understand the possible difficulties of applying RND to offline RL, we first reproduce the main experiment from \citet{rezaeifar2022offline}, which showed that RND is not discriminative enough to be used as a novelty bonus. For convenience, we provide the original figure from \citet{rezaeifar2022offline} in the \cref{app:prev-result}.
We also compare RND with a trained Q-ensemble (N = 25) from the SAC-N algorithm \citep{an2021uncertainty}. Similarly to \citet{rezaeifar2022offline}, we use simple state-action concatenation. Predictor and prior share the identical architecture of 4-layer MLPs. 

\begin{figure}[t]
    \vskip 0.2in
\centerline{\includegraphics[width=\columnwidth]{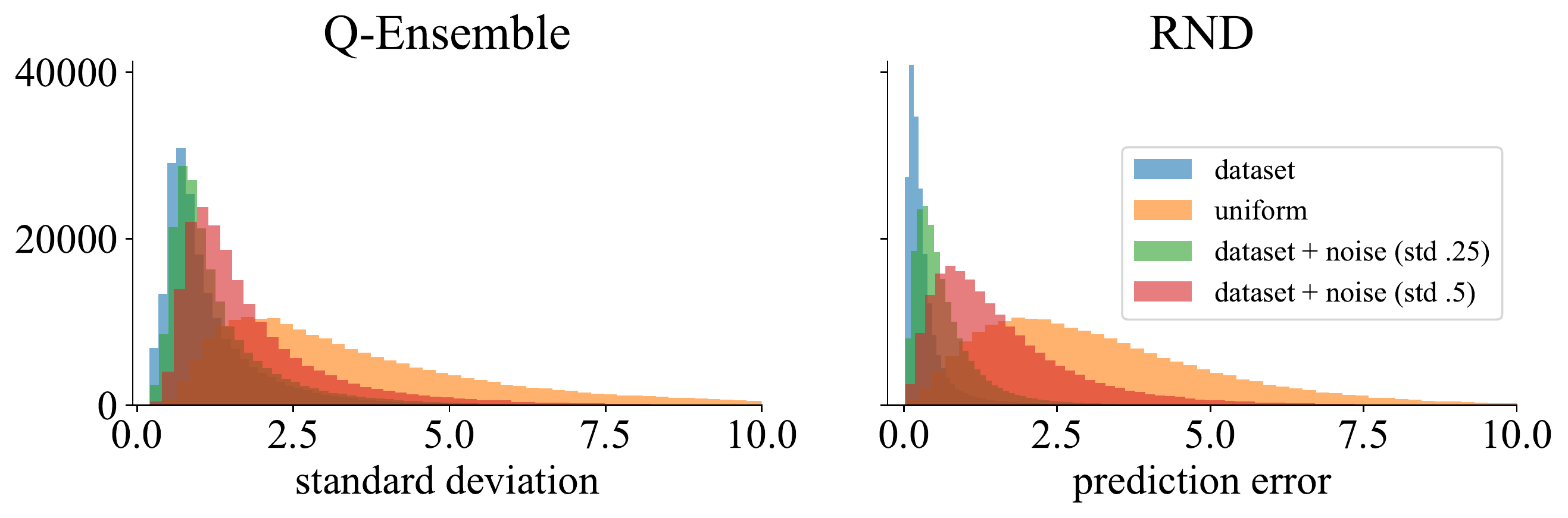}}
    \caption{Anti-exploration bonus \citep{rezaeifar2022offline} on the walker2d-medium dataset for trained SAC-N \cite{an2021uncertainty}, Q-ensemble (N = 25) and RND. Bonus is computed for state-action pairs from the original dataset and different perturbations of actions: random actions, dataset actions to which Gaussian noise is added with different scales. Both RND networks use simple state-action concatenation. The result is strikingly different from a similar figure in the \citet{rezaeifar2022offline} (we provide the original figure in the \cref{app:prev-result} for convenience). Contrary to previous research, it can be seen that RND is capable of distinguishing ID from OOD actions and is comparable to a trained Q-ensemble. 
    }
    \label{fig:disriminative}
    \vskip -0.2in
\end{figure}

The goal of the experiment (see \cref{fig:disriminative}) is to visually plot the anti-exploration bonus for ID state-action pairs and different perturbations of actions to model OOD data: random actions sampled from a uniform distribution and dataset actions to which Gaussian noise with different scales is added. 

To our surprise, the result on \cref{fig:disriminative} is strikingly different from previous work. It shows that RND is able to discriminate between ID and OOD actions with varying degrees of distributional shift and is comparable to a trained Q-ensemble. In contrast, \citet{rezaeifar2022offline} hypothesizes that RND can only work well out of the box for discrete action spaces and visual features, and concludes that extending it to continuous action spaces is not straightforward.

After further investigation of the open-sourced codebase\footnote{https://github.com/shidilrzf/Anti-exploration-RL} in search of discrepancies with our implementation, we found that the only difference is that, contrary to the advice of \citet{ciosek2019conservative}, \citet{rezaeifar2022offline} sets the predictor smaller than prior by two layers during RND pretraining. It is important to make the predictor larger or comparable in capacity to the prior so that it can minimize the loss to zero on the training dataset \citep{ciosek2019conservative}. However, the actual RND hyperparameters used in the final publication were not listed, so we cannot draw a definitive conclusion about the reason for such different results. 

\section{Concatenation Prior Hinders Bonus Minimization}
\label{demonstration}

\begin{figure*}[t]
    \begin{subfigure}[b]{0.333\textwidth}
        \centering
        \centerline{\includegraphics[width=\columnwidth]{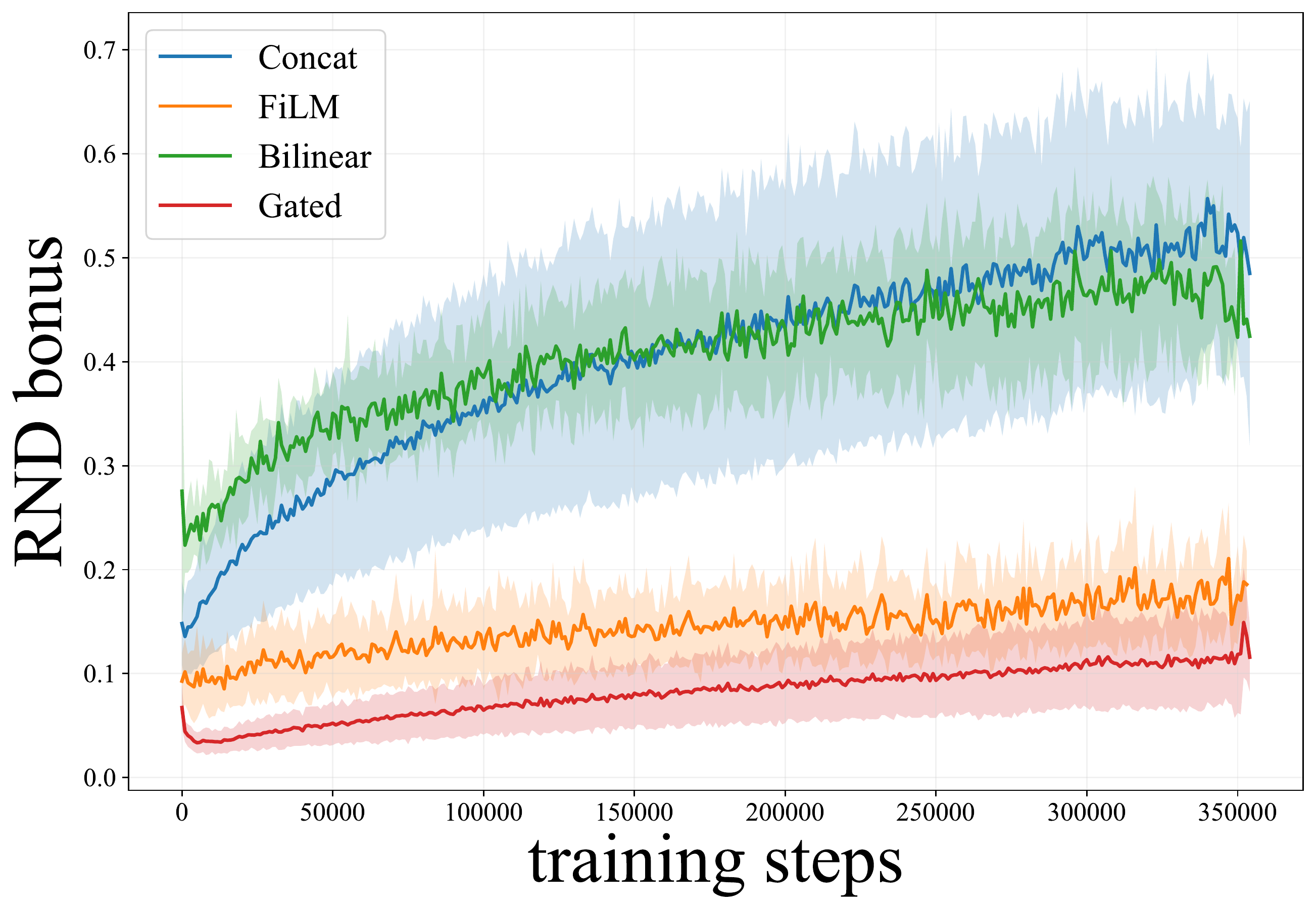}}
        \caption{RND bonus for dataset actions}
        \label{fig:concat_prior:rnd_data}
    \end{subfigure}
    \begin{subfigure}[b]{0.333\textwidth}
        \centering
        \centerline{\includegraphics[width=\columnwidth]{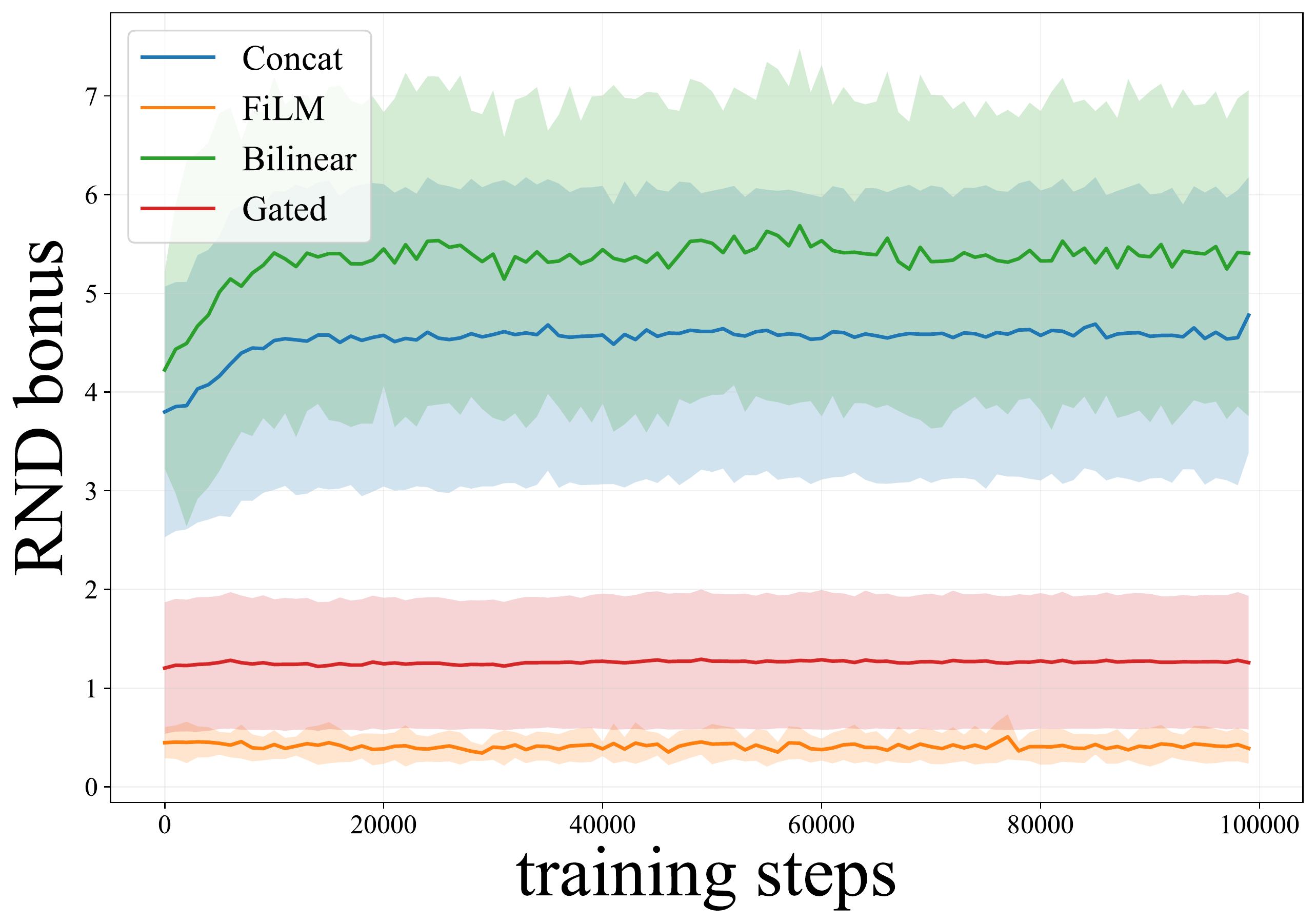}}
        \caption{RND bonus for actor actions}
        \label{fig:concat_prior:rnd_policy}
    \end{subfigure}
    \begin{subfigure}[b]{0.333\textwidth}
        \centering
        \centerline{\includegraphics[width=\columnwidth]{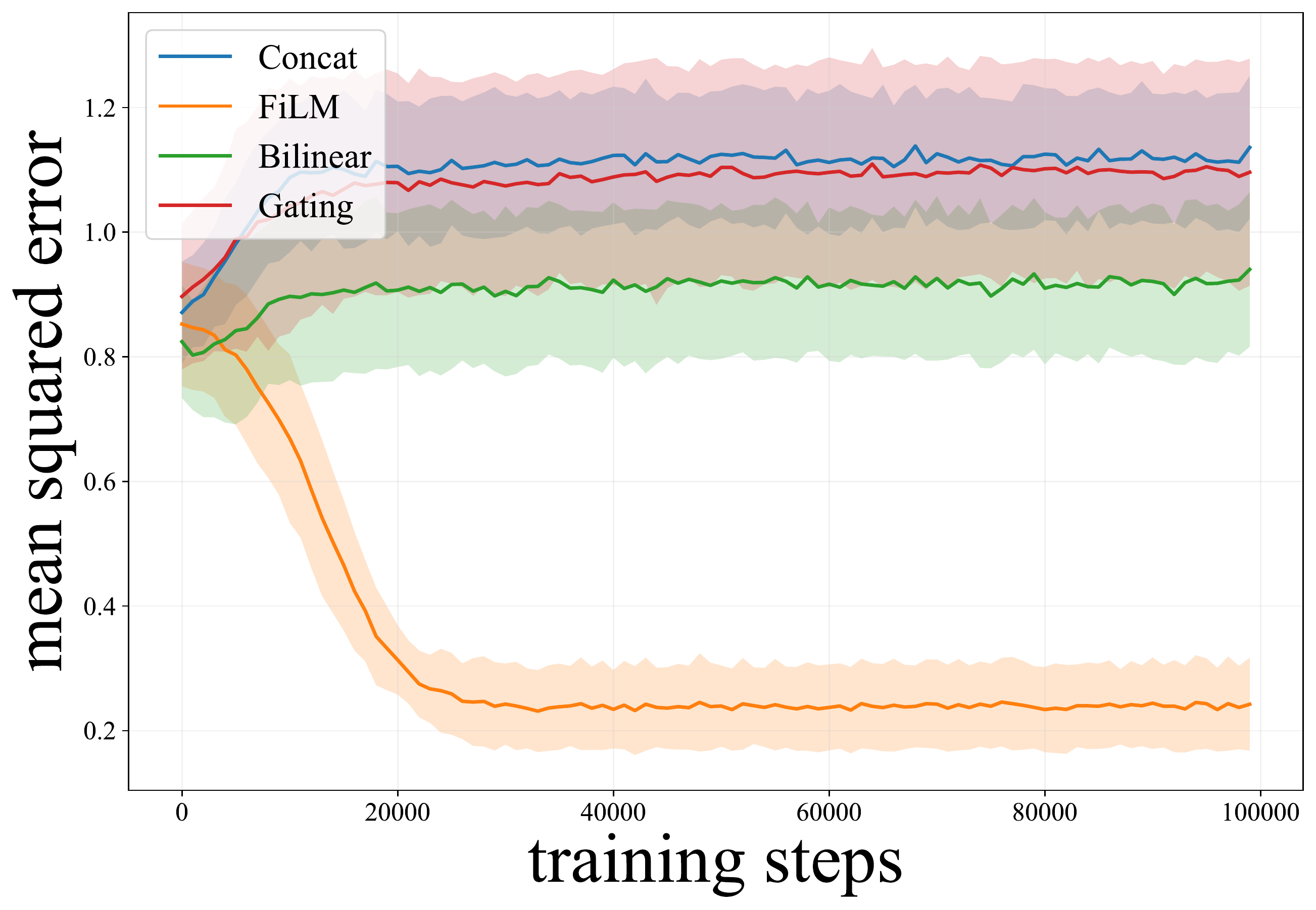}}
        \caption{Distance to dataset actions}
        \label{fig:concat_prior:act_mse}
    \end{subfigure}
    \caption{Effect of different state-action conditioning in the prior of RND on actor training. We use the halfcheetah, walker2d and hopper medium datasets, with 3 seeds each. For training procedure, see \cref{alg:bc-rnd} in the Appendix. \textbf{(a)} Anti-exploration bonus for in-dataset actions during RND pretraining. We additionally divide the bonus by the RND loss running standard deviation to increase its scale (see \cref{method}) so the anti-exploration bonus increases slightly over time as standard deviation decreases. However, this does not affect minimization by the actor and is needed to highlight the differences. \textbf{(b)} Anti-exploration bonus for actor actions during training. Ideally, it should converge to values close to the final values in (a). \textbf{(c)} Distance of actor actions to true in-dataset actions during training. Ideally, it should decrease, as actions closer to the behavioral policy have the lowest bonus by design.}
    \label{fig:concat_prior}
\end{figure*}

A well-behaved anti-exploration bonus for continuous action spaces, be it RND or any other, should satisfy at least two criteria. First, it should be discriminative enough to detect novel actions and downweight their value estimates (see \cref{eq:q-target}). Ideally, the bonus should be close to zero for ID data so that we do not bias the Q-function, as this can be detrimental to training. Second, it should allow the actor to easily minimize the bonus with gradient descent during training.

In \cref{reproduction}, we showed that RND can detect OOD actions. Nevertheless, naive use of RND as an anti-exploration bonus on top of the Soft Actor Critic algorithm \cite{haarnoja2018soft} still does not provide satisfactory performance (see \cref{fig:rnd-naive}) with scores lower than CQL \cite{kumar2020conservative} and SAC-N \cite{an2021uncertainty}. This gives us an hint that the problem may not be the discriminative power of RND, but that the actor cannot effectively minimize the anti-exploration bonus during training.

To test our hypothesis that the actor cannot effectively minimize the anti-exploration bonus, we further simplify the problem by removing the critic from the SAC algorithm but keeping the entropy bonus (see \cref{alg:bc-rnd} in the Appendix). We expect that, in such a setting, the actor will be able to successfully minimize the anti-exploration bonus to the possible minimum, i.e. comparable to the bonus for the ground truth data at the end of the RND pretraining. As a consequence, since dataset actions provide the minimum bonus by design, we also expect that the distance from the agent to dataset actions should be small.

We set predictor architecture to state-action concatenation. Additionally, we explore different conditioning schemes for the prior. We use the halfcheetah, walker2d and hopper medium datasets, with 3 seeds each. \cref{fig:concat_prior} compares the anti-exploration bonus for dataset actions during RND pretraining (see \cref{fig:concat_prior:rnd_data}) and for agent actions during training (see \cref{fig:concat_prior:rnd_policy}). 

As one can see for all prior architectures except one, the anti-exploration bonus during actor training is much higher than it should be according to the values on the dataset actions. These results confirm our hypothesis. Furthermore, we can note from \cref{fig:concat_prior:act_mse} that the actor cannot clone the behavioral policy, since the distance to the dataset actions can even increase during training. 

However, RND with the FiLM prior architecture allows the actor to effectively minimize the anti-exploration bonus and successfully clone the behavioral policy. This suggests that, with the right inductive bias for the prior, we can solve the problems of naive RND and possibly achieve better results.

\section{Anti-Exploration by Random Network Distillation}
\label{method}

We are now ready to present \textbf{SAC-RND}: a new offline RL method for continuous action spaces, based on our findings in \cref{reproduction} and \cref{demonstration}. It is simple, ensemble-free and achieves state-of-the-art results comparable to ensemble-based methods. We have chosen the Soft Actor-Critic \citep{haarnoja2018soft} algorithm as the backbone of the method. In this section, we will explain how the RND is trained and how we define the anti-exploration bonus. 

\begin{table}[t]
    \caption{Comparison of different RND predictors. Prior uses FiLM conditioning. Predictor uses conditioning in the first layer. All scores are averaged over 3 random seeds. Halfcheetah tasks are omitted, as we found them non-representative of the final performance on harder tasks. 
    }
    \label{method:predictors-table}
    \vskip 0.1in
    \begin{small}
        \begin{adjustbox}{max width=\columnwidth}
		\begin{tabular}{l|rrrr}
			\toprule
            Task Name & Concat & Gating & Bilinear & FiLM \\
			\midrule
			hopper-medium-v2          & 94.8  & 39.7 & 98.4   & 86.3 \\
			hopper-medium-expert-v2   & 71.5  & 59.3 & 110.3  & 102.7 \\
			hopper-medium-replay-v2   & 100.3 & 51.3 & 100.8  & 100.3 \\
			\midrule
			walker2d-medium-v2        & 94.8  & 82.3 & 92.8   & 95.1  \\
			walker2d-medium-expert-v2 & 86.1  & 84.2 & 108.9  & 110.0 \\
			walker2d-medium-replay-v2 & 90.3  & 87.5 & 88.3   & 75.7 \\
			\midrule
			Average                   & 89.6  & 67.3 & \textbf{99.9} & 95.0 \\
			\bottomrule
		\end{tabular}
        \end{adjustbox}
    \end{small}
    \vskip -0.1in
\end{table}

\textbf{Random Network Distillation}. We pretrain RND with MSE loss between prior and predictor embeddings, stopping gradient through prior and freezing both networks afterwards during SAC training. We keep both networks similar in size to the agent and critic, which are 4 layer MLPs. Contrary to \citet{burda2018exploration, ciosek2019conservative}, we do not add additional layers to the predictor to prevent undesirable results. This is because, when the predictor size is bigger than prior on state-based tasks (not image-based as in original work by \citet{burda2018exploration}), we observe that it can sometimes overgeneralize to OOD prior embeddings. 

According to \cref{demonstration}, for the prior, we use FiLM conditioning on penultimate layer before nonlinearity. In principle, the predictor can be arbitrary \cite{ciosek2019conservative}, but in practice, its architecture and conditioning type can also affect performance. We conduct a preliminary study on a small subset of the D4RL Gym tasks to select the best-performing conditioning. Based on the results in \cref{method:predictors-table}, we chose a predictor with bilinear conditioning in the first layer, as it showed the best performance.

\textbf{Anti-Exploration Bonus}. We define the anti-exploration bonus similarly to RND loss as 
\begin{equation}
\label{eq:offline-rnd-bonus}
 b(s, a) = \lVert f_{\psi}(s, a) - \bar{f}_{\bar{\psi}}(s, a) \rVert_{2} ^ 2
\end{equation}

and additionally divide it by RND loss running standard deviation (which is tracked during pretraining phase) to increase its scale uniformly among environments. Such scaling simplifies hyperparameter search, shrinking the possible range of useful $\alpha$ coefficients that control the level of conservatism during training.

For detailed training procedure and full SAC losses, we refer to \cref{alg:sac-rnd} in the Appendix (differences with the original SAC algorithm are highlighted in blue).

\section{Experiments}
\label{exp}

In this section, we present an empirical evaluation of our method using the D4RL benchmark on the Gym domain (\cref{exp:gym}) and the more challenging AntMaze domain (\cref{exp:antmaze}).
Next, we provide additional analysis and visual insight into why FiLM conditioning in the prior might be beneficial (\cref{exp:explanation}). Finally, we present an ablation that compares more variations of conditioning for predictor and prior (\cref{exp:cond-pairs}). For each experiment, we also list the exact hyperparameters in \cref{app:hyperparams} and implementation details in \cref{app:impl-details}. Additionally, we analyse sensitivity to hyperparameters in \cref{app:eop}. 

\subsection{Evaluation on the Gym Domain}
\label{exp:gym}

\textbf{Setup}. We evaluate our method on all available datasets for the HalfCheetah, Walker2d and Hopper tasks in the Gym domain of the D4RL benchmark. For ensemble-free baselines, we chose CQL \citep{kumar2020conservative}, IQL \citep{kostrikov2021offline}, TD3+BC \citep{fujimoto2021minimalist}, which show good results and are widely used in practice. We also report scores for vanilla SAC \citep{haarnoja2018soft}. For ensemble-based baselines, we chose SAC-N \& EDAC \citep{an2021uncertainty} and the more recent RORL \citep{yang2022rorl}, which currently achieve state-of-the-art scores in this domain. We follow the \citet{an2021uncertainty} and train for 3M gradient steps, evaluating on 10 episodes.

\textbf{Results}. The resulting scores are presented in \cref{exp:gym-table}. We see that SAC-RND stands out from the ensemble-free methods and outperforms them by a wide margin, achieving a mean score comparable to EDAC and only slightly behind RORL. Note that we do not use ensembles, whereas SAC-N can require up to 500 critics, EDAC up to 50 and RORL up to 20. In addition, we compare our proposed changes with the naive predictor and prior, confirming that our modifications are essential for achieving good performance (see \cref{fig:rnd-naive}).

\begin{table*}[t]
    \caption{SAC-RND evaluation on the Gym domain. We report the final normalized score averaged over 4 random seeds on v2 datasets. TD3 + BC and IQL scores are taken from \citet{lyu2022mildly}. CQL, SAC, SAC-N and EDAC scores are taken from \citet{an2021uncertainty}. RORL scores are taken from \citet{yang2022rorl}. 
    }
    \vskip 0.05in
    \label{exp:gym-table}
    \begin{small}
        \begin{adjustbox}{max width=\textwidth}
		\begin{tabular}{l|rrrr|rrr|r}
            \multicolumn{1}{c}{} & \multicolumn{4}{c}{Ensemble-free} & \multicolumn{3}{c}{Ensemble-based} \\
			\toprule
            \textbf{Task Name} & \textbf{SAC} & \textbf{TD3+BC} & \textbf{IQL} & \textbf{CQL} & \textbf{SAC-N} & \textbf{EDAC} & \textbf{RORL} & \textbf{SAC-RND} \\
            \midrule
            halfcheetah-random & 29.7 $\pm$ 1.4 & 11.0 $\pm$ 1.1 & 13.1 $\pm$ 1.3 & 31.1 $\pm$ 3.5 & 28.0 $\pm$ 0.9 & 28.4 $\pm$ 1.0 & 28.5 $\pm$ 0.8 & 29.0 $\pm$ 1.5 \\
            halfcheetah-medium & 55.2 $\pm$ 27.8 & 48.3 $\pm$ 0.3 & 47.4 $\pm$ 0.2 & 46.9 $\pm$ 0.4 & 67.5 $\pm$ 1.2 & 65.9 $\pm$ 0.6 & 66.8 $\pm$ 0.7 & 66.6 $\pm$ 1.6 \\
            halfcheetah-expert & -0.8 $\pm$ 1.8 & 96.7 $\pm$ 1.1 & 95.0 $\pm$ 0.5 & 97.3 $\pm$ 1.1 & 105.2 $\pm$ 2.6 & 106.8 $\pm$ 3.4 & 105.2 $\pm$ 0.7 & 105.8 $\pm$ 1.9 \\
            halfcheetah-medium-expert & 28.4 $\pm$ 19.4 & 90.7 $\pm$ 4.3 & 86.7 $\pm$ 5.3 & 95.0 $\pm$ 1.4 & 107.1 $\pm$ 2.0 & 106.3 $\pm$ 1.9 & 107.8 $\pm$ 1.1 & 107.6 $\pm$ 2.8 \\
            halfcheetah-medium-replay & 0.8 $\pm$ 1.0 & 44.6 $\pm$ 0.5 & 44.2 $\pm$ 1.2 & 45.3 $\pm$ 0.3 & 63.9 $\pm$ 0.8 & 61.3 $\pm$ 1.9 & 61.9 $\pm$ 1.5 & 54.9 $\pm$ 0.6 \\
            halfcheetah-full-replay & 86.8 $\pm$ 1.0 & - & - & 76.9 $\pm$ 0.9 & 84.5 $\pm$ 1.2 & 84.6 $\pm$ 0.9 & - & 82.7 $\pm$ 0.9 \\
            \midrule
            hopper-random & 9.9 $\pm$ 1.5 & 8.5 $\pm$ 0.6 & 7.9 $\pm$ 0.2 & 5.3 $\pm$ 0.6 & 31.3 $\pm$ 0.0 & 25.3 $\pm$ 10.4 & 31.4 $\pm$ 0.1 & 31.3 $\pm$ 0.1 \\
            hopper-medium & 0.8 $\pm$ 0.0 & 59.3 $\pm$ 4.2 & 66.2 $\pm$ 5.7 & 61.9 $\pm$ 6.4 & 100.3 $\pm$ 0.3 & 101.6 $\pm$ 0.6 & 104.8 $\pm$ 0.1 & 97.8 $\pm$ 2.3 \\
            hopper-expert & 0.7 $\pm$ 0.0 & 107.8 $\pm$ 7.0 & 109.4 $\pm$ 0.5 & 106.5 $\pm$ 9.1 & 110.3 $\pm$ 0.3 & 110.1 $\pm$ 0.1 & 112.8 $\pm$ 0.2 & 109.7 $\pm$ 0.3 \\
            hopper-medium-expert & 0.7 $\pm$ 0.0 & 98.0 $\pm$ 9.4 & 91.5 $\pm$ 14.3 & 96.9 $\pm$ 15.1 & 110.1 $\pm$ 0.3 & 110.7 $\pm$ 0.1 & 112.7 $\pm$ 0.2 & 109.8 $\pm$ 0.6 \\
            hopper-medium-replay & 7.4 $\pm$ 0.5 & 60.9 $\pm$ 18.8 & 94.7 $\pm$ 8.6 & 86.3 $\pm$ 7.3 & 101.8 $\pm$ 0.5 & 101.0 $\pm$ 0.5 & 102.8 $\pm$ 0.5 & 100.5 $\pm$ 1.0 \\
            hopper-full-replay & 41.1 $\pm$ 17.9 & - & - & 101.9 $\pm$ 0.6 & 102.9 $\pm$ 0.3 & 105.4 $\pm$ 0.7 & - & 107.3 $\pm$ 0.1 \\
            \midrule
            walker2d-random & 0.9 $\pm$ 0.8 & 1.6 $\pm$ 1.7 & 5.4 $\pm$ 1.2 & 5.1 $\pm$ 1.7 & 21.7 $\pm$ 0.0 & 16.6 $\pm$ 7.0 & 21.4 $\pm$ 0.2 & 21.5 $\pm$ 0.1 \\
            walker2d-medium & -0.3 $\pm$ 0.2 & 83.7 $\pm$ 2.1 & 78.3 $\pm$ 8.7 & 79.5 $\pm$ 3.2 & 87.9 $\pm$ 0.2 & 92.5 $\pm$ 0.8 & 102.4 $\pm$ 1.4 & 91.6 $\pm$ 2.8 \\
            walker2d-expert & 0.7 $\pm$ 0.3 & 110.2 $\pm$ 0.3 & 109.9 $\pm$ 1.2 & 109.3 $\pm$ 0.1 & 107.4 $\pm$ 2.4 & 115.1 $\pm$ 1.9 & 115.4 $\pm$ 0.5 & 114.3 $\pm$ 0.6 \\
            walker2d-medium-expert & 1.9 $\pm$ 3.9 & 110.1 $\pm$ 0.5 & 109.6 $\pm$ 1.0 & 109.1 $\pm$ 0.2 & 116.7 $\pm$ 0.4 & 114.7 $\pm$ 0.9 & 121.2 $\pm$ 1.5 & 105.0 $\pm$ 7.9 \\
            walker2d-medium-replay & -0.4 $\pm$ 0.3 & 81.8 $\pm$ 5.5 & 73.8 $\pm$ 7.1 & 76.8 $\pm$ 10.0 & 78.7 $\pm$ 0.7 & 87.1 $\pm$ 2.4 & 90.4 $\pm$ 0.5 & 88.7 $\pm$ 7.7 \\
            walker2d-full-replay & 27.9 $\pm$ 47.3 & - & - & 94.2 $\pm$ 1.9 & 94.6 $\pm$ 0.5 & 99.8 $\pm$ 0.7 & - & 109.2 $\pm$ 1.8 \\
            \midrule
            Average & 16.2 & 67.5 & 68.9 & 73.6 & 84.4 & \textbf{85.2} & \textbf{85.7} & \textbf{85.2} \\
			\bottomrule
		\end{tabular}
        \end{adjustbox}
    \end{small}
    \vskip -0.1in
\end{table*}

\subsection{Evaluation on the AntMaze Domain}
\label{exp:antmaze}

\textbf{Setup}. We evaluate our method on all datasets available for the AntMaze domain of the D4RL benchmark. Ensemble-free baselines are the same as in \cref{exp:gym}. For ensemble-based baselines, we chose RORL \citep{yang2022rorl} and MSG \citep{ghasemipour2022so}, the latter of which, to our knowledge, currently has the best mean score for this domain. We do not include SAC-N and EDAC, as there are no public results for them on this domain, and we were also unable to obtain a non-zero result. We follow the \citet{an2021uncertainty} and train for 3M gradient steps, evaluating on 100 episodes.

\textbf{Results}. The resulting scores are presented in \cref{exp:antmaze-table}. \citet{kostrikov2021offline} has shown that many offline RL methods that perform well on the Gym domain fail on the AntMaze domain.
It can be seen that, on the AntMaze domain, SAC-RND shows good results that are on par with ensembles, and outperforms ensemble-free methods. This also shows that our choice of predictor and prior generalises well to new domains. Note that, in addition to ensembles, both MSG and RORL require pre-training or supervision with behavioural cloning in order to achieve reported results, while our method does not require any additional modifications.

\begin{table}[h]
    \caption{SAC-RND evaluation on AntMaze domain. We report the final normalized score averaged over 4 random seeds on v1 datasets. IQL, CQL, MSG scores are taken from \citet{ghasemipour2022so} and SAC from \cite{kumar2020conservative}. TD3+BC, RORL scores are taken from \citet{yang2022rorl}.
    }
    \label{exp:antmaze-table}
    \begin{small}
        \begin{adjustbox}{max width=\columnwidth}
		\begin{tabular}{l|rrrr|rr|r}
            \multicolumn{1}{c}{} & \multicolumn{4}{c}{Ensemble-free} & \multicolumn{2}{c}{Ensemble-based} \\
			\toprule
            \textbf{Task Name} & \textbf{SAC} & \textbf{TD3+BC} & \textbf{IQL} & \textbf{CQL} & \textbf{RORL} & \textbf{MSG} & \textbf{SAC-RND} \\
            \midrule
            antmaze-umaze & 0.0 & 78.6  & 87.5  & 74.0 & 97.7 $\pm$ 1.9 & 97.8 $\pm$ 1.2 & 97.2 $\pm$ 1.2 \\
            antmaze-umaze-diverse & 0.0 & 71.4  & 62.2  & 84.0 & 90.7 $\pm$ 2.9 & 81.8 $\pm$ 3.0 & 83.5 $\pm$ 7.7 \\
            antmaze-medium-play & 0.0 & 10.6  & 71.2  & 61.2 & 76.3 $\pm$ 2.5 & 89.6 $\pm$ 2.2 & 65.5 $\pm$ 35.7 \\
            antmaze-medium-diverse & 0.0 & 3.0  & 70.0  & 53.7  & 69.3 $\pm$ 3.3 & 88.6 $\pm$ 2.6 & 88.5 $\pm$ 9.2 \\
            antmaze-large-play & 0.0 & 0.2  & 39.6  & 15.8  & 16.3 $\pm$ 11.1 & 72.6 $\pm$ 7.0 & 67.2 $\pm$ 6.1 \\
            antmaze-large-diverse & 0.0 & 0.0  & 47.5  & 14.9  & 41.0 $\pm$ 10.7 & 71.4 $\pm$ 12.2 & 57.6 $\pm$ 22.7 \\
            \midrule
            Average & 0.0 & 27.3 & 63.0 & 50.6 & 65.2 & \textbf{83.6} & \textbf{76.6} \\
			\bottomrule
		\end{tabular}
        \end{adjustbox}
    \end{small}
    \vskip -0.1in
\end{table}

\subsection{Why is FiLM Conditioning Beneficial for Bonus Minimization?}
\label{exp:explanation}

\begin{figure*}[t]
    \begin{center}
    \begin{subfigure}[b]{0.245\textwidth}
        \centering
        \centerline{\includegraphics[width=\columnwidth]{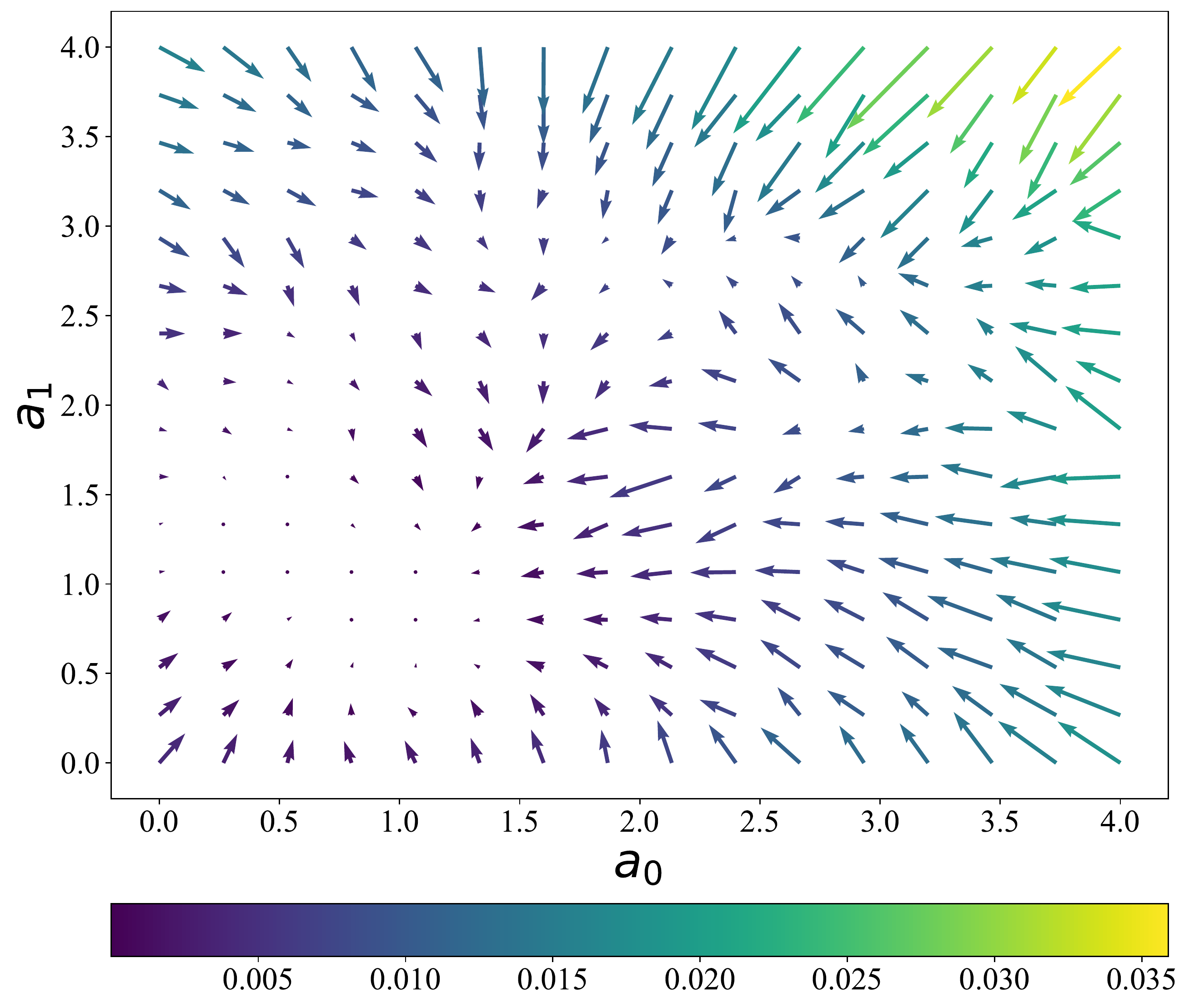}}
    \end{subfigure}
    \vspace{0.3cm}
    \begin{subfigure}[b]{0.245\textwidth}
        \centering
        \centerline{\includegraphics[width=\columnwidth]{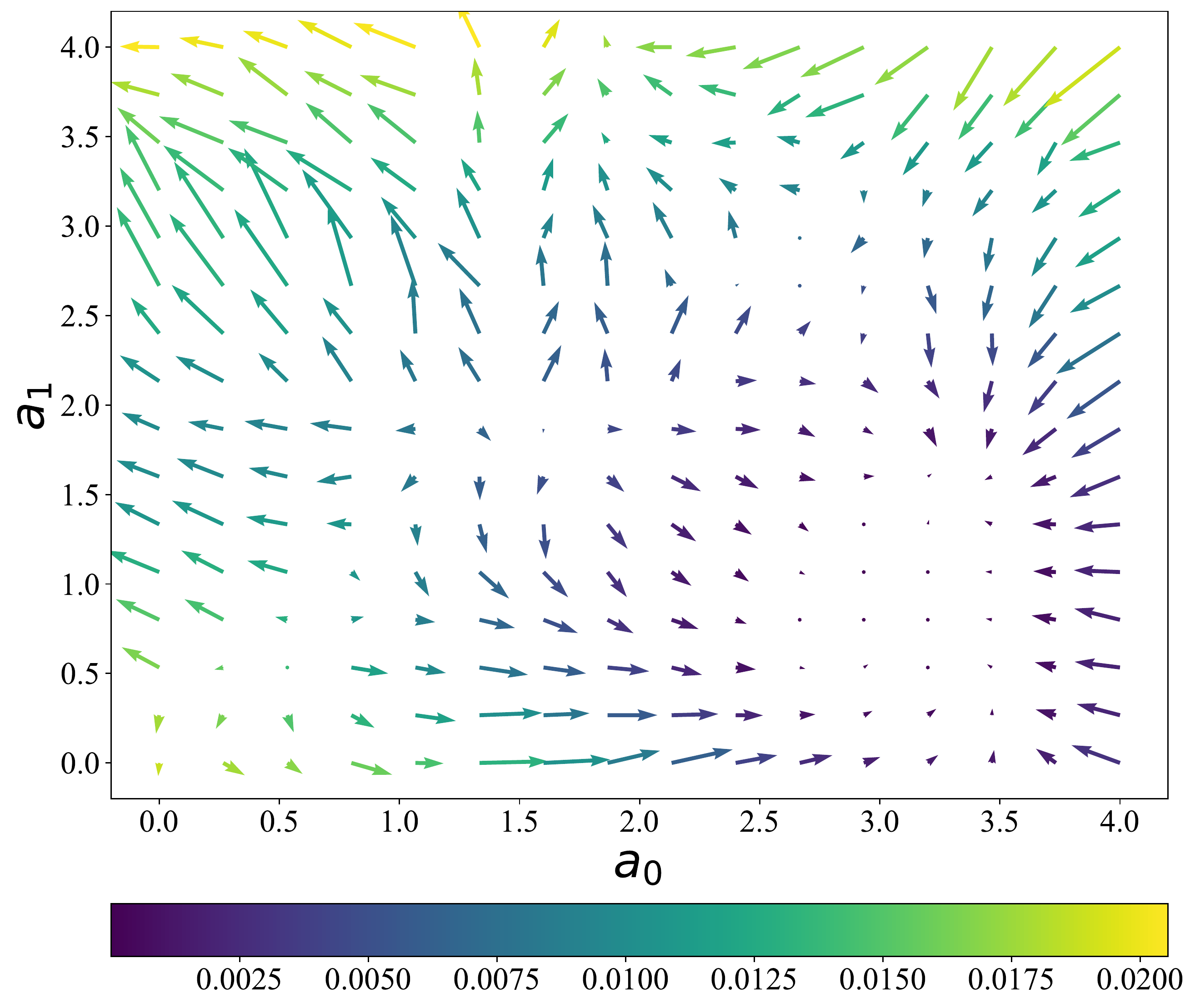}}
    \end{subfigure}
    \begin{subfigure}[b]{0.245\textwidth}
        \centering
        \centerline{\includegraphics[width=\columnwidth]{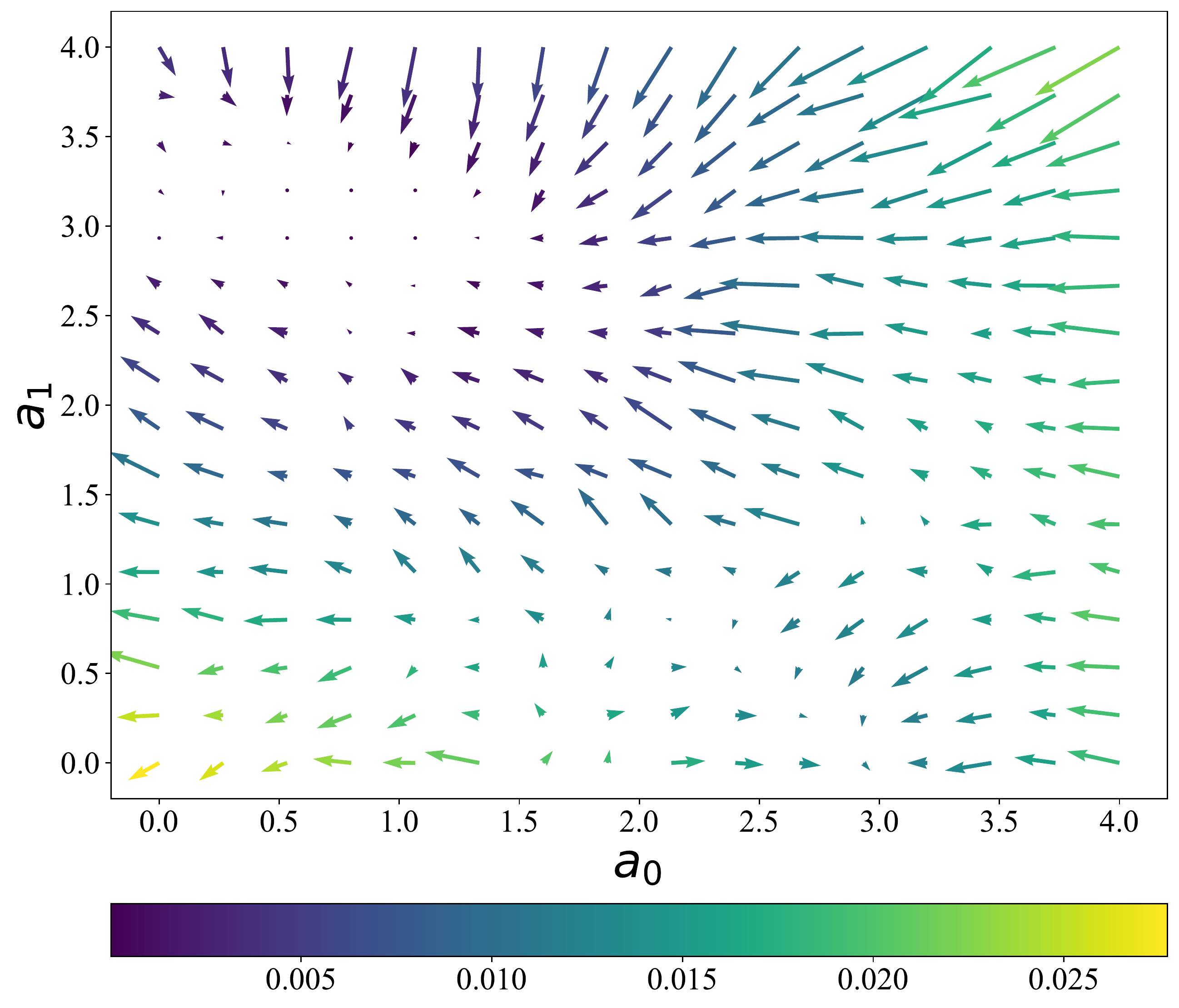}}
    \end{subfigure}
    \begin{subfigure}[b]{0.245\textwidth}
        \centering
        \centerline{\includegraphics[width=\columnwidth]{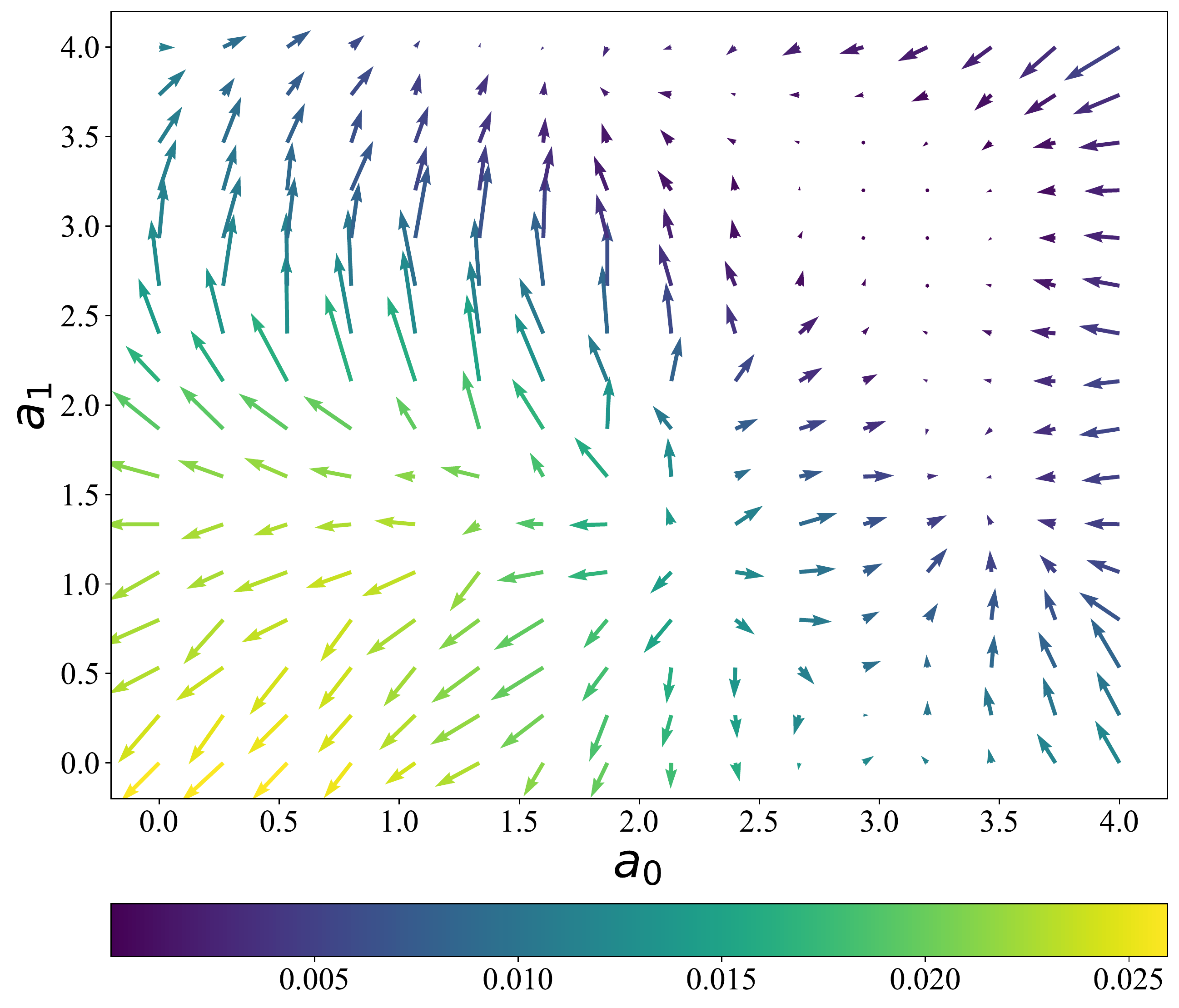}}
    \end{subfigure}
    
    \begin{subfigure}[b]{0.245\textwidth}
        \centering
        \centerline{\includegraphics[width=\columnwidth]{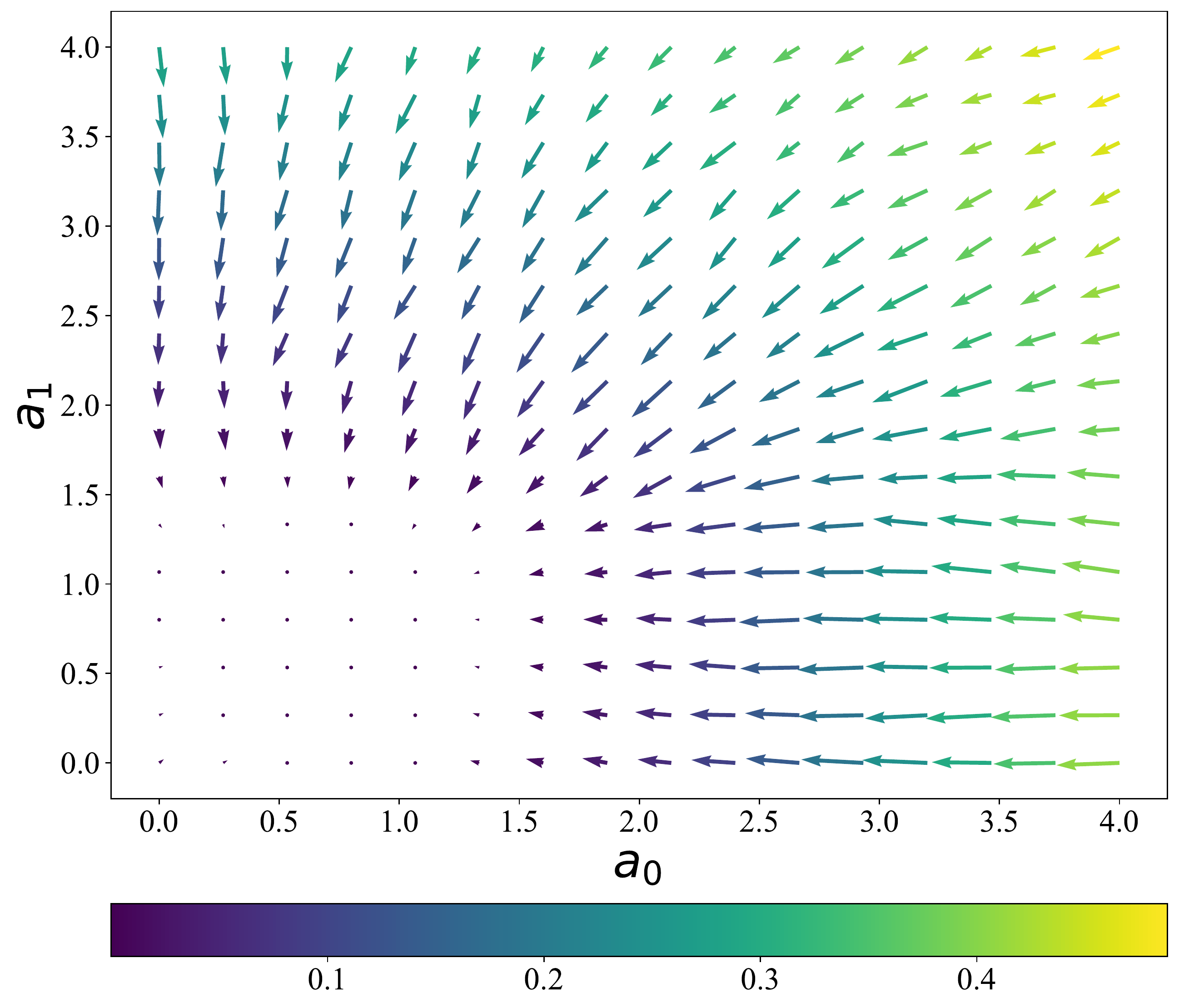}}
        \caption{State 0}
    \end{subfigure}
    \begin{subfigure}[b]{0.245\textwidth}
        \centering
        \centerline{\includegraphics[width=\columnwidth]{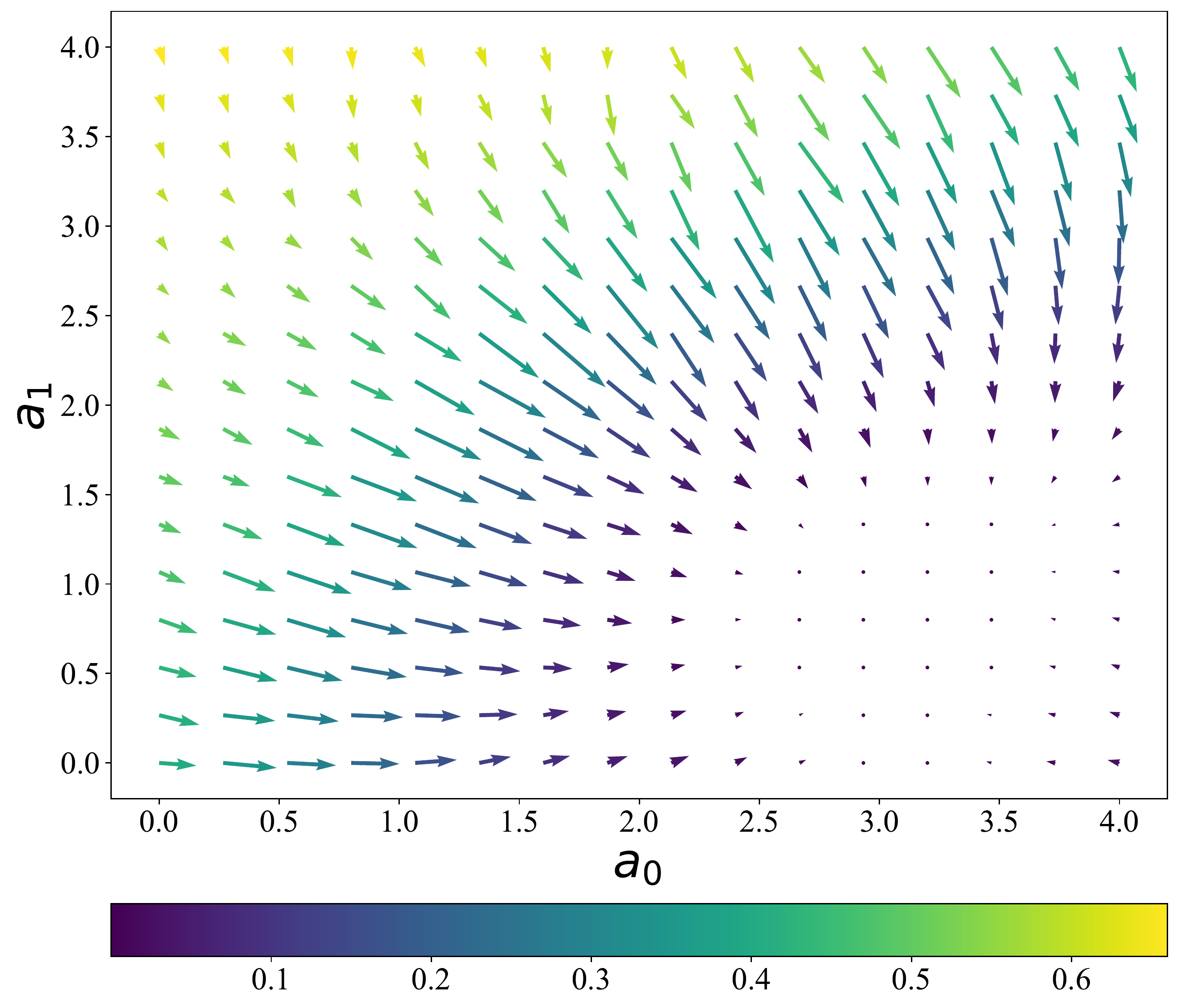}}
        \caption{State 1}
    \end{subfigure}
    \begin{subfigure}[b]{0.245\textwidth}
        \centering
        \centerline{\includegraphics[width=\columnwidth]{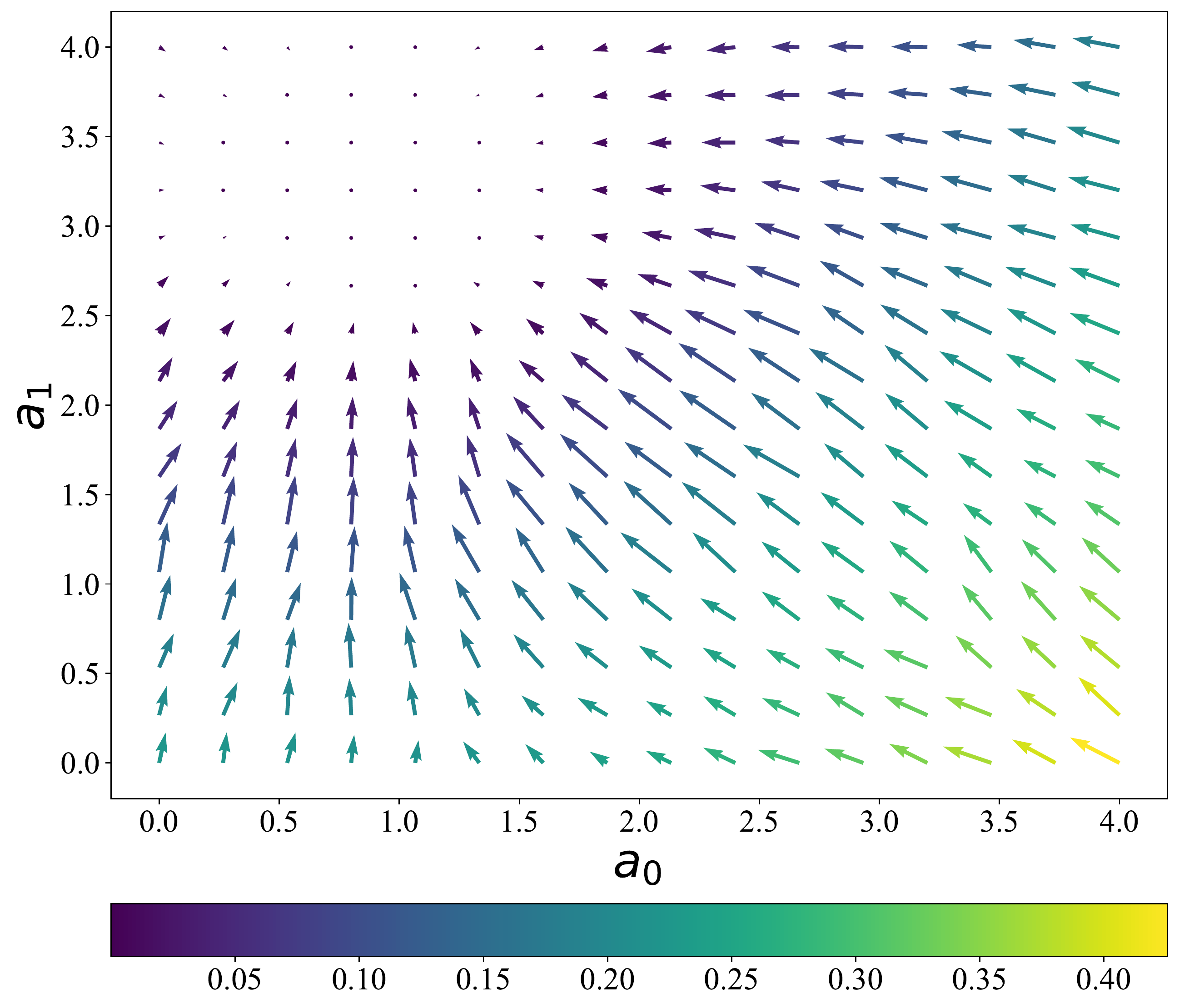}}
        \caption{State 2}
    \end{subfigure}
    \begin{subfigure}[b]{0.245\textwidth}
        \centering
        \centerline{\includegraphics[width=\columnwidth]{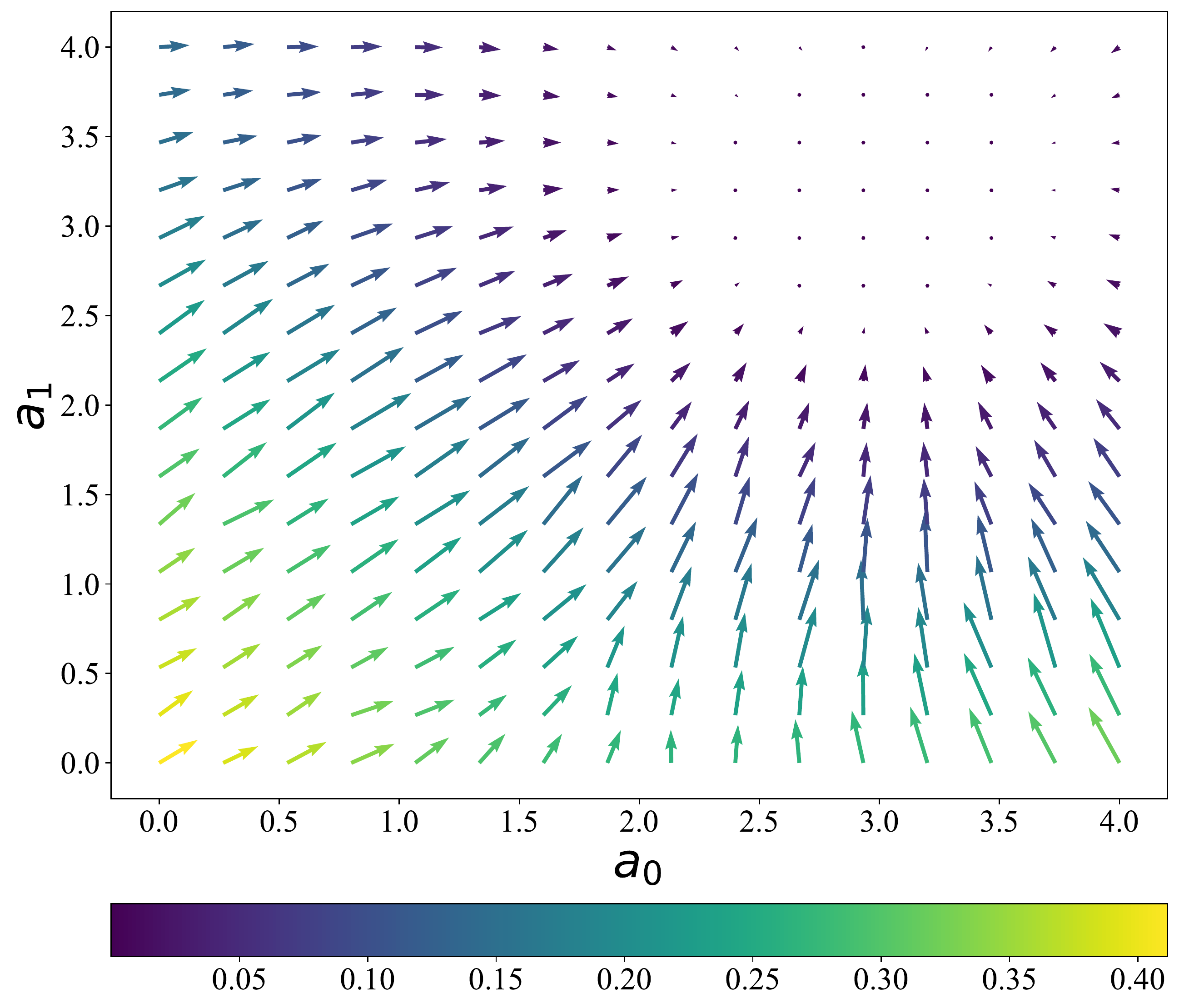}}
        \caption{State 3}
    \end{subfigure}
    \end{center}
    
    \caption{Actions' anti-gradient field for the anti-exploration bonus conditioned on four categorical states at each corner for the toy problem introduced in \cref{exp:explanation}. We visualize the dataset in \cref{fig:toy-dataset} in the appendix. The top row corresponds to RND with concatenation conditioning in the prior, while the bottom row corresponds to FiLM conditioning. As can be seen, the resulting anti-gradients for concatenation are noisy, while the directions for FiLM are smooth over the entire available action space.}
    \label{fig:gradient_field}
\end{figure*}

In \cref{demonstration}, we showed that FiLM conditioning in the RND prior significantly improved the actors' ability to minimize the anti-exploration bonus. Since the issue occurred during actor training, we hypothesize that this may be related to the anti-exploration bonus optimization landscape. In this section, we analyze the anti-gradient fields for conditioning with concatenation or FiLM for the prior network.

For the purpose of analysis, we design a toy dataset with only four categorical states and two-dimensional actions sampled uniformly in each corner of the grid (see \cref{app:toy-dataset} for dataset visualization and generation details). 

We fix the hyperparameters and pretrain two RNDs that differ only in the type of prior conditioning. The predictor uses simple concatenation. Next, in \cref{fig:gradient_field}, we plot the two-dimensional anti-gradient field for the anti-exploration bonus conditioned on each state. The effect of FiLM becomes more apparent in these plots. While the resulting anti-gradients for concatenation  are noisy and only point in the direction of the minimum in a small neighbourhood, the directions for FiLM are smooth over the entire available action space and point to the correct global minimum for each state. While we cannot draw general conclusions from such a demonstration, based on the results of \cref{demonstration}, we hypothesize that a similar phenomenon might exist in high-dimensional problems as well.

\subsection{Exploring More Conditioning Pairs}
\label{exp:cond-pairs}

One might wonder (1) how different types of conditioning for predictor and prior interact with each other and (2) where to introduce conditioning in terms of depth for it to be most beneficial. 

\begin{figure}[h]
\centerline{\includegraphics[width=0.8\columnwidth]{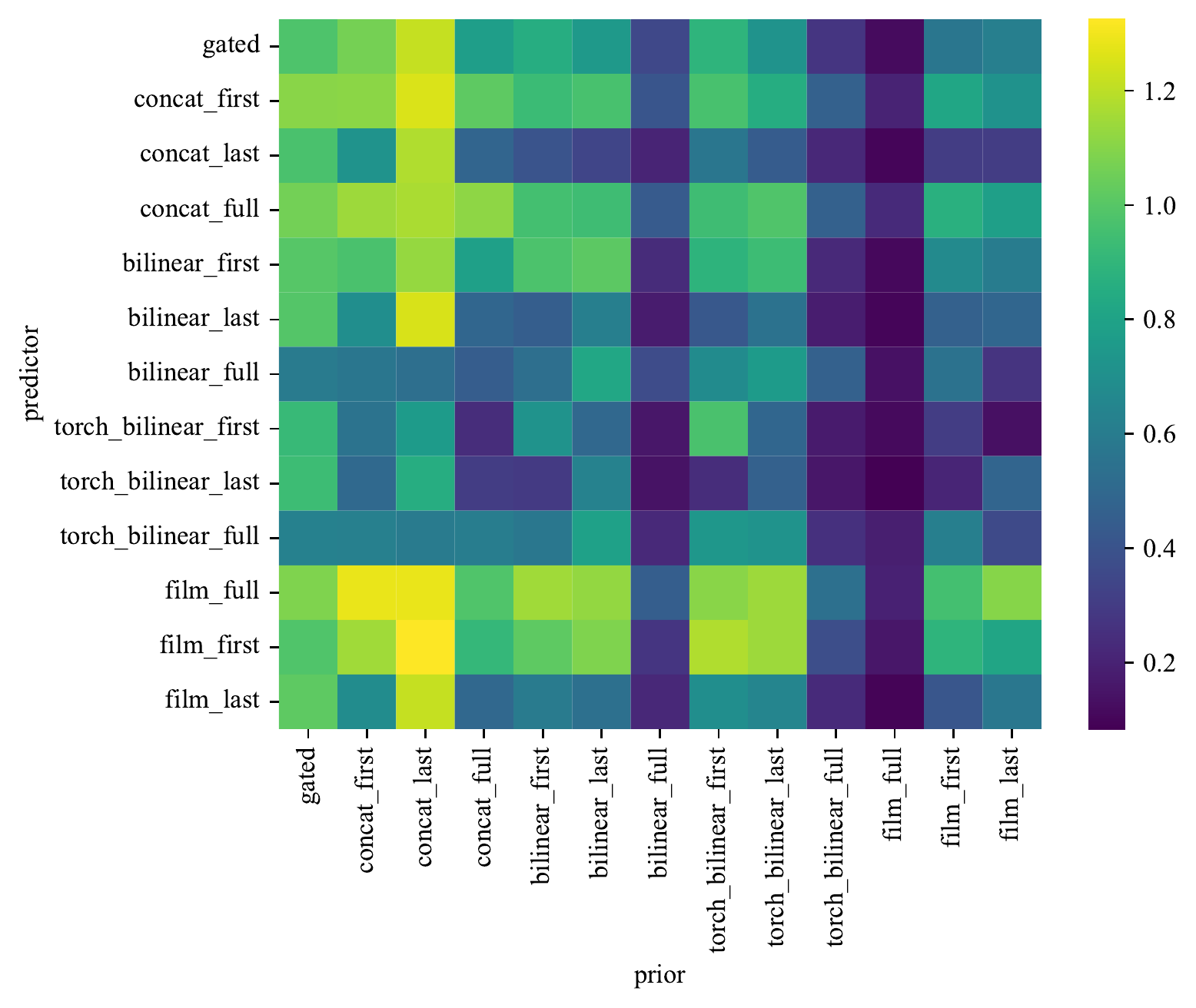}}
    \caption{Mean squared error between actions of the actor trained with different conditioning for the predictor \& prior and actions of the behavioral policy. We use the halfcheetah, walker2d and hopper medium datasets, with 3 seeds each. It can be seen that conditioning on each layer is beneficial for the priors, while for the predictors, it is better to condition on the last layer. Note that this experiment is in the setting of \cref{demonstration}, that is, without a critic.}
    \label{fig:more-priors}
    \vskip -0.2in
\end{figure}

To answer these questions, we return to the experiment from \cref{demonstration} and generate more variations for each type (where it is possible): conditioning on the first layer, on the last layer, and on all layers. We also look at two variations of the bilinear layer: full, as presented in \citet{jayakumar2020multiplicative}, and simplified, which is used by default in PyTorch. In \cref{fig:more-priors} we plot the final MSE between the resulting policy and the behavioural one on the training data. Two interesting observations can be made from these findings. 

First, FiLM may not be the only architecture with the right inductive biases for the prior, and both bilinear types with conditioning on all layers can also achieve similar results. However, compared to FiLM, inner bilinear layers are much more computationally expensive, as they involve at least one 3D weight tensor and two additional 2D weight tensors, and the hidden dimensions are usually much higher than the input dimensions. 

Second, it appears that conditioning on the last layer is most beneficial for the predictor, while conditioning on all layers is beneficial for the prior. In spite of that, it is difficult to draw broad conclusions, as different types may work well for new problems and domains.

\section{Related Work}
\label{related}
 
\textbf{Model-free offline RL}. 
Most offline RL approaches focus on the distribution shift problem and overestimation bias of Q-values for OOD actions. Some researchers address this by imposing strict constraints for policy updates, penalizing the divergence from the behavioral policy with KL divergence, maximum mean discrepancy (MMD) distance \citep{kumar2019stabilizing, wu2019behavior}, simple mean squared error (MSE) \citep{fujimoto2021minimalist}, or by re-weighting behavioral policy actions with the estimated advantages \citep{nair2020awac}. Others directly regularize Q-values by lowering return estimates for OOD actions, preventing overestimation for unseen actions. For instance, \citet{kumar2020conservative}, \citet{ghasemipour2022so} and \citet{rezaeifar2022offline} explicitly introduce an optimization term that lowers Q-values for OOD actions, while \citet{an2021uncertainty} penalizes implicitly by utilizing the lower-confidence bound (LCB) of Q-values. Alternatively, the evaluation of OOD actions can be avoided altogether by using the upper expectile value function \citep{kostrikov2021offline} or by policy optimization within a latent action space \citep{chen2022latent, zhou2021plas, akimovlet}. 

In our work, we follow the anti-exploration approach \citep{rezaeifar2022offline}. In contrast to \citet{an2021uncertainty, ghasemipour2022so, yang2022rorl}, we completely eliminate ensembles for uncertainty estimation, thus reducing computational overhead without sacrificing performance. Moreover, unlike \citet{rezaeifar2022offline}, we have succeeded in using an RND for novelty detection in offline RL for continuous action spaces.

\textbf{Estimation bias in Q-learning}. 
In both offline and online reinforcement learning, off-policy Q-learning methods suffer from an overestimation bias in the temporal difference learning target \citep{van2016deep, fujimoto2018addressing}. This phenomenon is orthogonal to overestimation due to unseen actions in offline RL, as it occurs even in the presence of strong conservatism constraints. It is mainly caused by target prediction errors for seen transitions and their propagation due to the maximum operation $max_{a' \in A}Q(s', a')$. To address this problem, \citet{fujimoto2018addressing} introduced clipped double Q learning \citep{van2016deep} in TD3, which uses a minimum of two critics. This approach was later used by \citet{haarnoja2018soft} in SAC to improve stability and accelerate convergence. 


In our work, we use clipped double Q-learning \cite{fujimoto2018addressing}, since SAC-RND is based on SAC \citep{haarnoja2018soft}, and found it beneficial for stability. However, to ensure that it does not introduce additional conservatism, which can be a confounding factor for the impact of RND, we always set the number of critics to two.

\textbf{Uncertainty estimation in offline RL}. 
Uncertainty estimation is a popular technique in reinforcement learning and is used for a variety of purposes such as exploration, planning, and robustness. In offline RL, its use is mostly limited to modeling epistemic uncertainty \citep{clements2019estimating}, including measuring the prediction confidence of dynamics models \citep{yu2020mopo, kidambi2020morel} or critics \citep{an2021uncertainty, rezaeifar2022offline}. This approach can be further used to induce uncertainty-aware penalization during training.

Alternatively, uncertainty can help overcome suboptimal conservatism by designing more flexible offline approaches, e.g., conditioning on different levels of confidence to dynamically adjust the level of conservatism during evaluation \citep{hong2022confidence} or using Bayesian perspective to design an optimal adaptive offline RL policy \citep{ghosh2022offline}.

In our work, we estimate epistemic uncertainty and use it as an anti-exploration bonus to induce conservatism. Unlike previous approaches, we do not use ensembles to estimate epistemic uncertainty.

\textbf{Efficient ensembles} Ensembles are a powerful and simple non-Bayesian baseline for uncertainty estimation that outperform Bayesian neural networks in practice \citep{lakshminarayanan2017simple}. However, training deep ensembles can be both memory intensive and computationally demanding, making the design of efficient ensembles an attractive research direction for which numerous methods have been developed. For example, \citet{gal2016dropout} proposed to use dropout to approximate Bayesian inference in deep Gaussian processes, and \citet{durasov2021masksembles} derived a method to interpolate between dropout and full ensembles with fixed masks and controllable overlap between them. Meanwhile, \citet{wen2020batchensemble} significantly reduced the cost by defining each weight matrix as the Hadamard product of a shared weight among all ensemble members and a rank-one matrix per member. 

Recently, \citet{ghasemipour2022so} showed that, in offline RL, none of the most popular approaches for efficient ensembles are capable of delivering performance that is comparable to naive ensembles, and that more work is needed in this research direction. In our work, we chose an alternative path for uncertainty estimation with RND, which was shown to a fast and competitive counterpart to ensembles \cite{ciosek2019conservative}.

\section{Conclusion} 

In this work, we revisited the results from previous research \citep{rezaeifar2022offline}, showing that with a naive choice of conditioning for the RND prior, it becomes infeasible for the actor to effectively minimize the anti-exploration bonus and discriminativity is not an issue. To solve this, we proposed conditioning based on FiLM, which led us to a new ensemble-free method called SAC-RND. We empirically validated that it achieves results comparable to ensemble-based methods and outperforms its ensemble-free counterparts. As such, we believe that our work is a valuable contribution to anti-exploration and uncertainty estimation in offline RL. In addition, we further outline the possible limitations of this work in the \cref{app:limit}. 

\bibliography{sac_rnd}
\bibliographystyle{icml2023}

\newpage
\appendix
\onecolumn

\section{Limitations}
\label{app:limit}

Several limitations of this work should be noted. While we provide extensive experiments and ablations on the well-established D4RL benchmark for continuous control, we lack the benchmarks for discrete actions and visual state spaces. Thus, we cannot be absolutely certain that the same pair of predictor and prior will generalize to the new domains, and additional exploration on new problems may be needed. However, the general properties of the good RND prior for offline RL uncovered in this work remain the same and should guide practitioners in the new applications of our method. Furthermore, we have explored only a limited set of the many available conditioning variations. Finally, our method inherits the limitations of anti-exploration style algorithms \cite{rezaeifar2022offline} --- sensitivity to the reward and RND bonus scale, as this greatly affects the level of conservatism, thus requiring the $\alpha$ sweep for each problem separately. To alleviate this limitation, we additionally explicitly checked the sensitivity to this parameter using the expected online performance metric in the \cref{app:eop}.

\section{Previous Research Results}
\label{app:prev-result}

\begin{figure}[h]
    \vskip 0.2in
\centerline{\includegraphics[width=\textwidth]{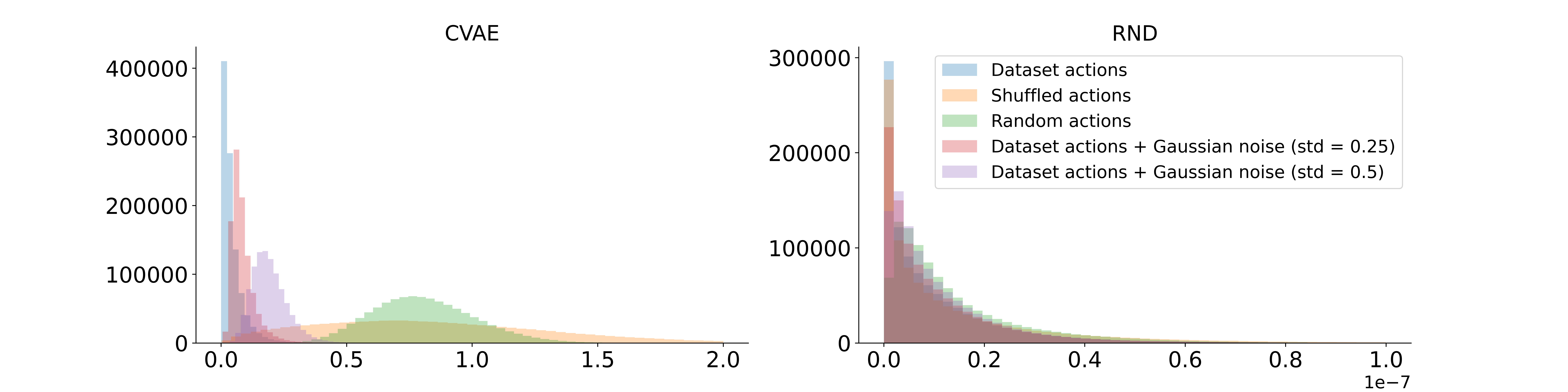}}
    \caption{Anti-exploration bonus on walker2d-medium dataset for RND and CVAE. Note that figure taken from \citet{rezaeifar2022offline} for a convenient comparison with our results in \cref{fig:disriminative}.
    }
    \label{fig:prev-research}
    \vskip -0.2in
\end{figure}

\section{Toy Dataset}
\label{app:toy-dataset}

\begin{figure}[h]
    \vskip 0.2in
\centerline{\includegraphics[width=.3\textwidth]{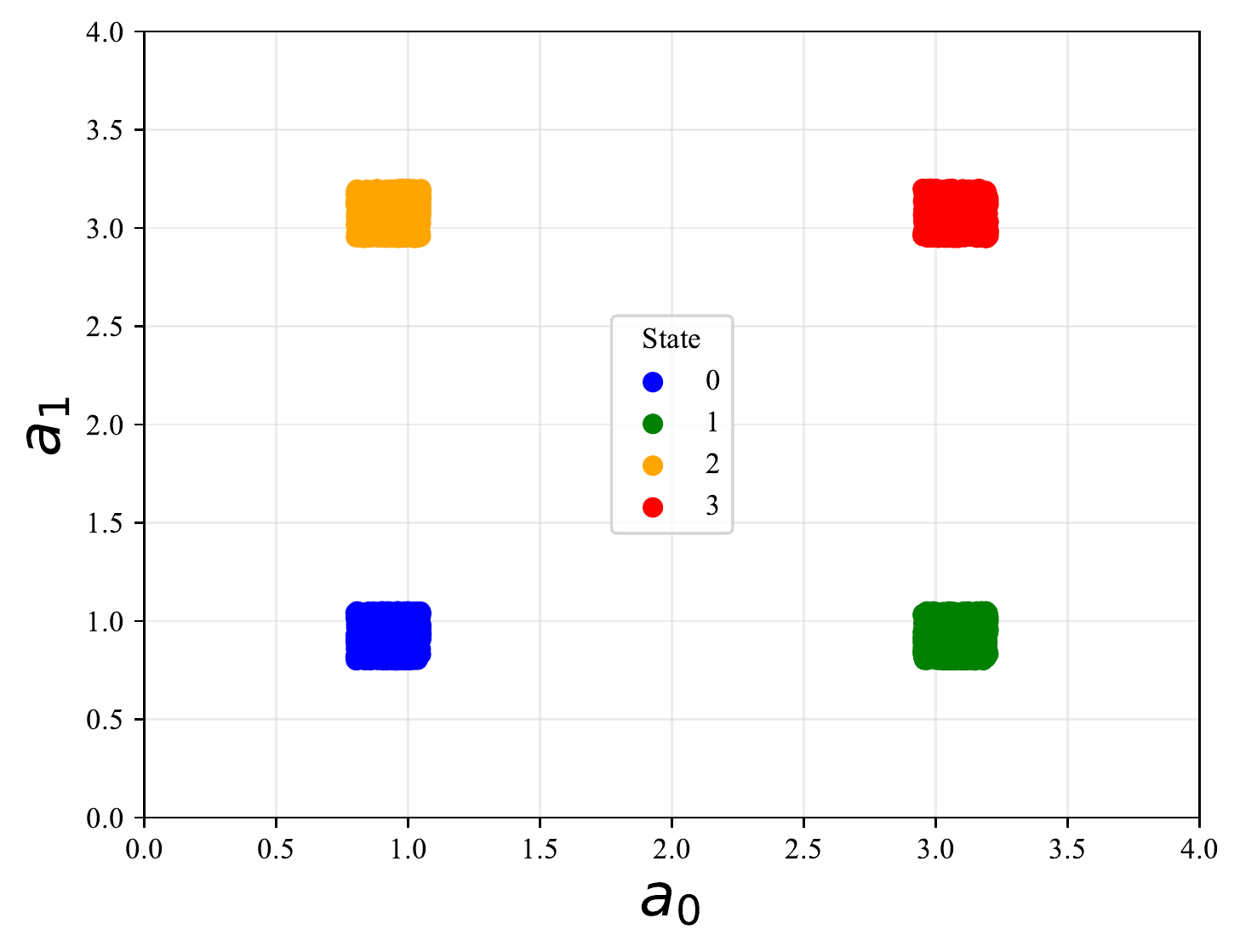}}
    \caption{Toy dataset visualization introduced in \cref{exp:explanation}. This toy dataset consists of four categorical states for each corner of the limited 2D actions grid. For each state, we uniformly sample 4096 two-dimensional actions within a limited square. We use one-hot encoding for the states during RND training.}
    \label{fig:toy-dataset}
    \vskip -0.2in
\end{figure}

\section{Implementation Details}
\label{app:impl-details}

 In our experiments, we use hyperparameters from \cref{app:shared-hps} where possible and sweep over $\alpha$ to pick the best value for each dataset. We implement all of our models using the Jax \citep{jax2018github} framework. For the exact implementation of conditioning variants for predictor and prior networks, refer to our code at \url{https://github.com/tinkoff-ai/ sac-rnd}. Similarly to \citet{nikulin2022q, kumar2022offline, smith2022walk}, we add Layer Normalization \citep{ba2016layer} to the critic after each layer as it greatly improves stability and convergence speed. For SAC-N in \cref{demonstration} we use the implementation from the CORL library \citep{tarasov2022corl}. All experiments were performed on V100 and A100 GPUs. With our implementation, each training for 3 million training steps usually takes $\sim40$ minutes to run  ($\sim15$ minutes for the typical 1 million steps).

\textbf{Gym Domain}. We use the v2 version of each dataset. We follow the \citet{an2021uncertainty} approach and run our algorithms for 3 million training steps and report the final normalized average score over 10 evaluation episodes. For the final experiments, we use 4 seeds, while using less for hyperparameter tuning. We tune the $\alpha$ coefficient over the $\{1.0, 3.0, 4.0, 5.0, 8.0, 9.0, 10.0, 13.0, 15.0, 20.0, 25.0\}$ range for the walker and hopper datasets. We found that the halfcheetah datasets require a lower level of conservatism, which is why we tune over the $\{0.001, 0.1, 0.3, 0.5, 0.8, 1.0, 2.0, 3.0, 4.0, 5.0, 6.0\}$ range for these datasets while keeping the same number of candidates. We follow the \citet{ghasemipour2022so} approach and choose the best $\alpha$ for each dataset (see \cref{app:gym-hps}).

\textbf{AntMaze Domain}. We use the v1 version of each dataset due to the fact that the v0 version has major problems and bugs during generation (e.g., some trajectories have discontinuities where the agent teleports from one part of the maze to another \footnote{https://github.com/Farama-Foundation/D4RL/issues/77}). We follow the \citet{an2021uncertainty} approach and run our algorithms for 3 million training steps and report the final normalized average score over 100 evaluation episodes. Same as \citet{chen2022latent}, we scale the reward by 100.0. We found that actor and critic require different levels of conservatism in these tasks, which is why we chose to decouple $\alpha$ and use separate values (the same approach was used in \citet{rezaeifar2022offline}). We tune the $\alpha$ for the actor in the $\{0.5, 1.0, 1.5\}$ range, and $\alpha$ for the critic in the $\{0.001, 0.01, 0.1\}$ range.  We follow the \citet{ghasemipour2022so} approach and choose the best $\alpha$ for each dataset (see \cref{app:antmaze-hps}).

\section{Hyperparameters}
\label{app:hyperparams}
\vskip -0.2in
\begin{table}[h]
    \caption{SAC-RND general hyperparameters.}
    \label{app:shared-hps}
    \vskip 0.1in
    \begin{center}
    \begin{small}
		\begin{tabular}{l|l}
			\toprule
            \textbf{Parameter} & \textbf{Value} \\
            \midrule
            optimizer                         & Adam~\citep{kingma2014adam} \\
            batch size                        & 1024 (256 on antmaze-*) \\
            learning rate (all networks)       & 1e-3 (3e-4 on antmaze-*) \\
            tau ($\tau$)                      & 5e-3 \\
            hidden dim (all networks)         & 256 \\
            num layers (all networks)         & 4 \\
            RND embedding dim (all tasks)     & 32 \\
            target entropy                    & -action\_dim \\
            gamma ($\gamma$)                  & 0.99 (0.999 on antmaze-*) \\
            nonlinearity                      & ReLU \\
   \bottomrule
		\end{tabular}
    \end{small}
    \end{center}
    \vskip -0.2in
\end{table}

\begin{table}[h]
    \caption{SAC-RND best hyperparameters used in D4RL Gym domain.}
    \label{app:gym-hps}
    \vskip 0.1in
    \begin{center}
    \begin{small}
    \begin{adjustbox}{max width=\textwidth}
		\begin{tabular}{l|r}
			\toprule
            \textbf{Task Name} & \textbf{$\alpha$} \\
            \midrule
            halfcheetah-random & 0.1 \\
            halfcheetah-medium & 0.3 \\
            halfcheetah-expert & 6.0 \\
            halfcheetah-medium-expert & 0.1 \\
            halfcheetah-medium-replay & 0.1 \\
            halfcheetah-full-replay & 3.0 \\
            \midrule
            hopper-random & 5.0 \\
            hopper-medium & 25.0 \\
            hopper-expert & 20.0 \\
            hopper-medium-expert & 15.0 \\
            hopper-medium-replay & 8.0 \\
            hopper-full-replay & 3.0 \\
            \midrule
            walker2d-random & 1.0 \\
            walker2d-medium & 8.0 \\
            walker2d-expert & 4.0 \\
            walker2d-medium-expert & 25.0 \\
            walker2d-medium-replay & 8.0 \\
            walker2d-full-replay & 3.0 \\
			\bottomrule
		\end{tabular}
    \end{adjustbox}
    \end{small}
    \end{center}
    \vskip -0.2in
\end{table}

\begin{table}[h!]
    \caption{SAC-RND best hyperparameters used in D4RL AntMaze domain.}
    \label{app:antmaze-hps}
    \vskip 0.1in
    \begin{center}
    \begin{small}
		\begin{tabular}{l|r|r}
			\toprule
            \textbf{Task Name} & $\alpha$ (actor) & $\alpha$ (critic) \\
            \midrule
            antmaze-umaze          & 1.0 & 0.1 \\
            antmaze-umaze-diverse  & 1.0 & 0.1 \\
            antmaze-medium-play    & 0.5 & 0.001 \\
            antmaze-medium-diverse & 1.0 & 0.01 \\
            antmaze-large-play     & 1.0 & 0.01 \\
            antmaze-large-diverse  & 0.5 & 0.01 \\
			\bottomrule
		\end{tabular}
    \end{small}
    \end{center}
    \vskip -0.1in
\end{table}

\section{Runtime Comparison}
\begin{table}[H]
    \caption{Runtime comparison for different algorithms on halfcheetah-medium-v2 dataset. We measure time on V100 GPU for standard 1M updates with batch size 256 and identical network sizes. For SAC-N \citep{an2021uncertainty} we report times for all ensemble size configurations from the original publication. Note that, as we discussed in \cref{app:impl-details}, all algorithms were implemented in Jax \citep{jax2018github}, which is typically much faster for small networks than PyTorch \citep{paszke2019pytorch} alternatives.}
    \label{app:antmaze-hps}
    \vskip 0.1in
    \begin{center}
    \begin{small}
		\begin{tabular}{l|r|r}
			\toprule
            \textbf{Algorithm} & \textbf{Updates / s} $\uparrow$ & \textbf{Runtime (m)} $\downarrow$ \\
            \midrule
            TD3+BC  & 1485 & 11.2   \\  
            SAC-2   & 1285 & 12.9   \\  
            SAC-RND & 850  & 19.6    \\  
            SAC-10  & 809  & 20.5    \\  
            SAC-20  & 559  & 29.7    \\  
            SAC-100 & 171  & 97.3    \\  
            SAC-200 & 93   & 178.3    \\  
            SAC-500 & 39   & 424.0    \\  
			\bottomrule
		\end{tabular}
    \end{small}
    \end{center}
    \vskip -0.1in
\end{table}

\section{Sensitivty to Hyperparameters}
\label{app:eop}

\begin{figure}[H]
    \vskip 0.2in
\centerline{\includegraphics[width=0.4\textwidth]{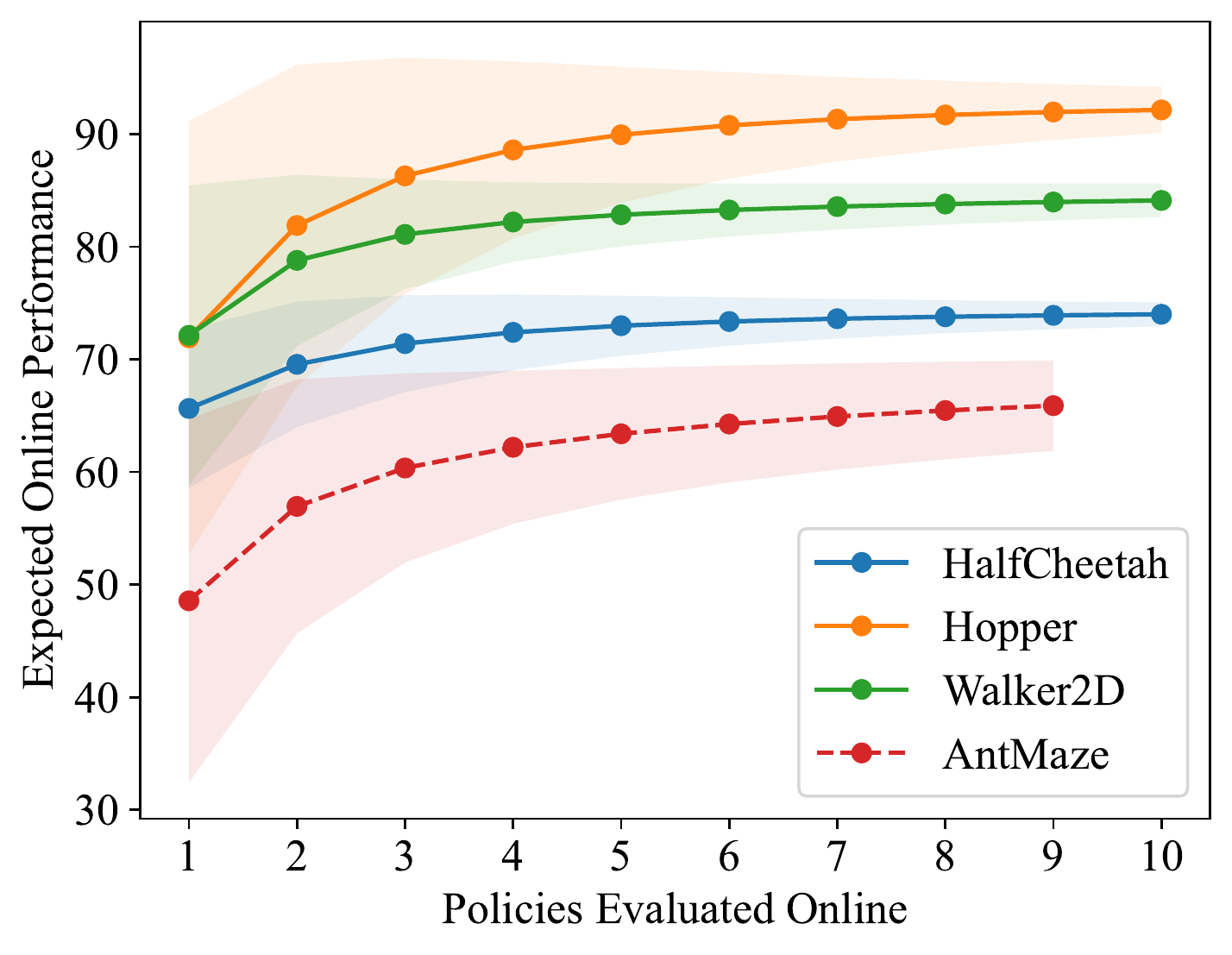}}
    \caption{Expected Online Performance \citep{kurenkov2022showing} under uniform offline policy selection. It can be seen, that for satisfactory results in all domains a budget of at least five policies for online evaluations is needed.}
    \label{fig:eop}
    \vskip -0.2in
\end{figure}

\newpage
\section{Architecture Visualization}
\begin{figure*}[h]
    \begin{center}
    \begin{subfigure}[t]{0.49\textwidth}
        \centering
        \centerline{\includegraphics[width=\columnwidth]{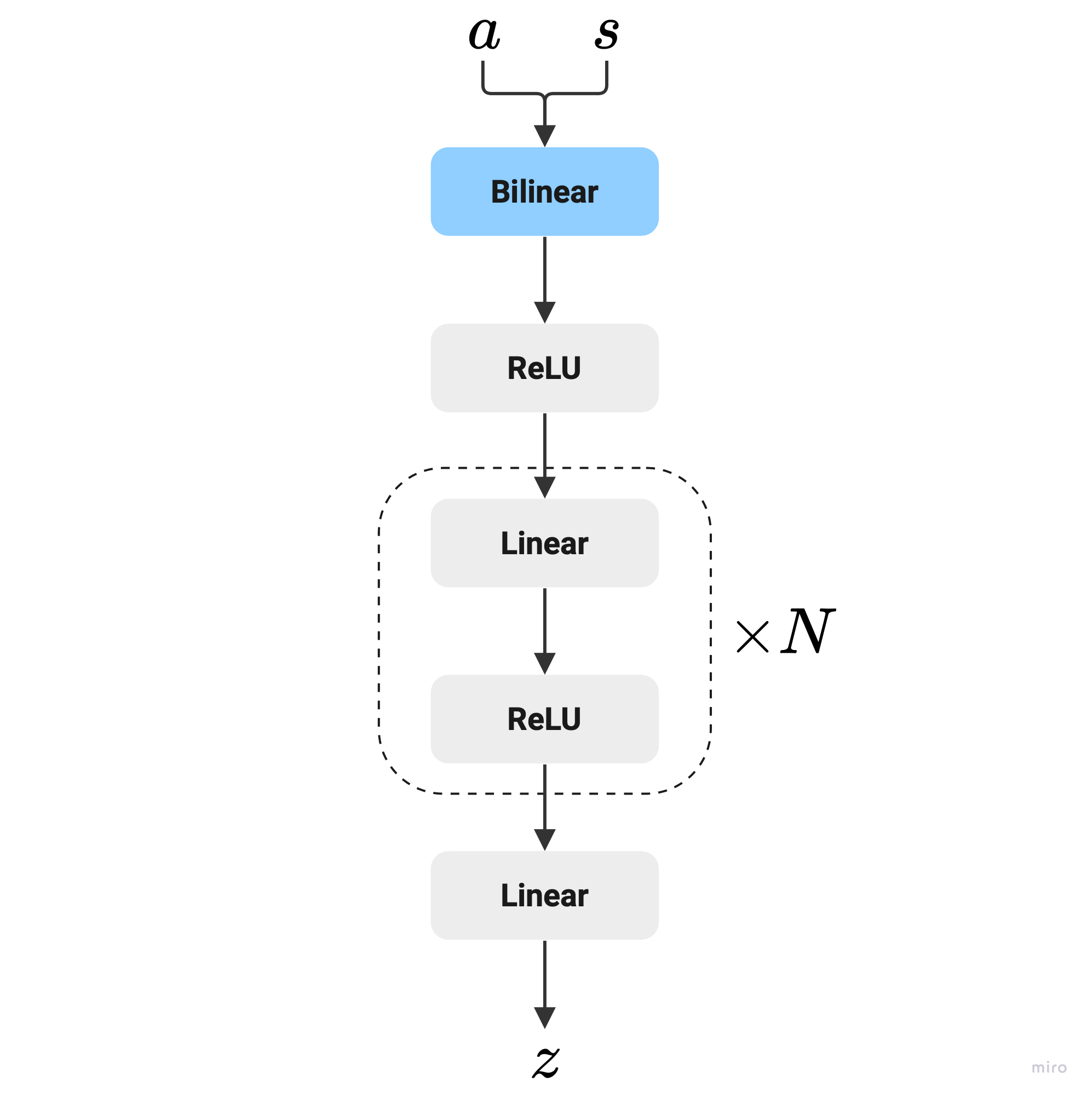}}
        \caption{Predictor}
    \end{subfigure}
    \begin{subfigure}[t]{0.49\textwidth}
        \centering
        \centerline{\includegraphics[width=\columnwidth]{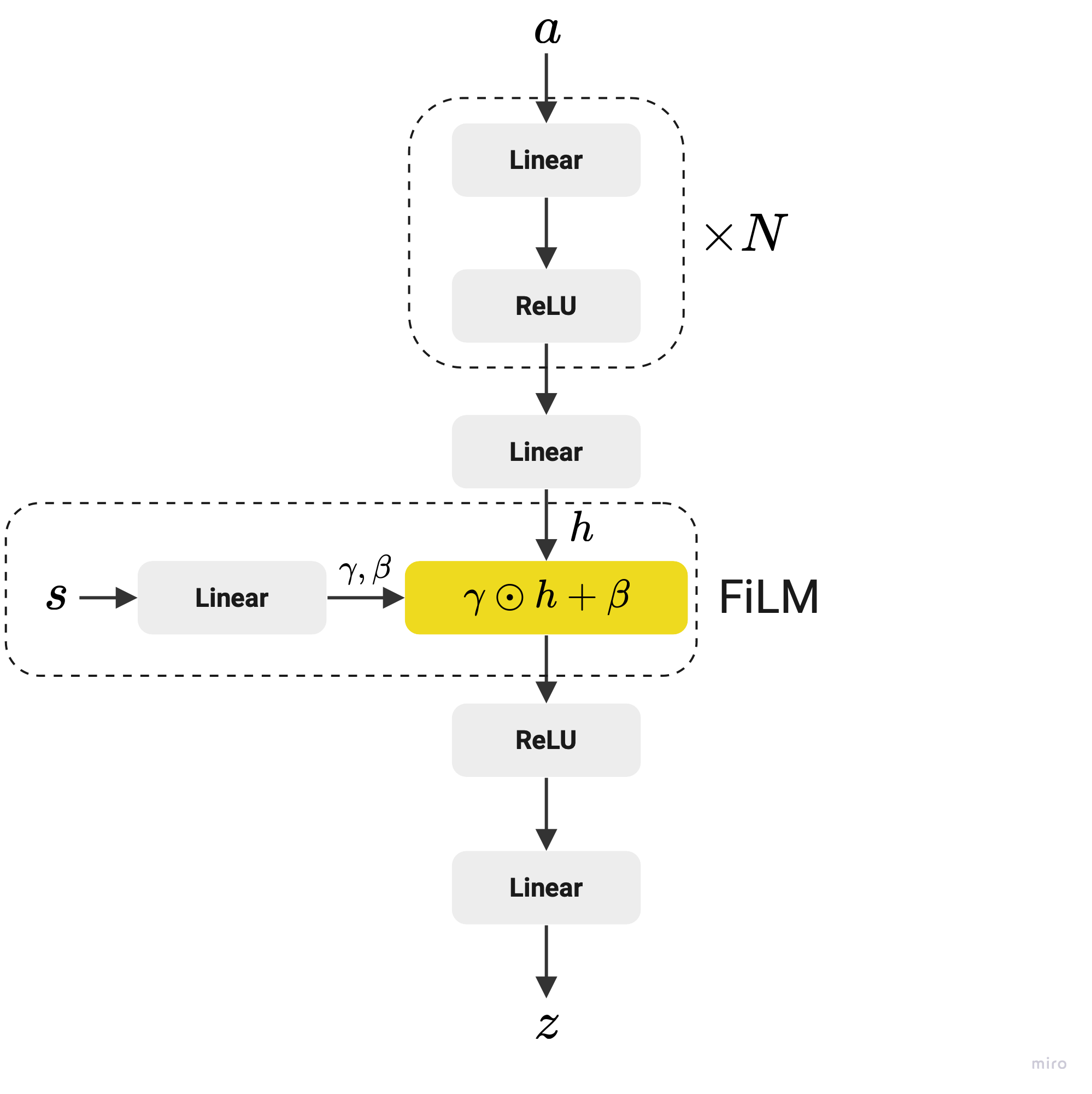}}
        \caption{Prior}
    \end{subfigure}

    \end{center}
    \caption{Visualization of the final RND architecture described in the \autoref{method}. For the predictor we use bilinear \cite{jayakumar2020multiplicative} conditioning on the first layer. For the prior we use FiLM \cite{perez2018film} conditioning on the penultimate layer before nonlinearity. States are encoded with a single linear layer of double hidden size, which output is then divided into equal parts for the $\gamma$ and $\beta$. While the number of linear layers can be arbitrary, in our work we use $N = 2$, so the network size is 4 layers in total (see \autoref{app:shared-hps}).}
    \label{fig:rnd_diagram}
\end{figure*}

\newpage
\section{Pseudocode}

\begin{algorithm*}[h]
   \caption{Soft Actor-Critic with Random Network Distillation (SAC-RND)}
   \label{alg:sac-rnd}
\begin{algorithmic}
    \STATE Initialize policy parameters $\theta$, Double Q-function parameters $\{ \phi_{1}, \phi_{2} \}$, \textcolor{blue}{RND predictor and prior parameters $\{\psi, \psi'\}$}, and offline replay buffer $\mathcal{D}$ 
    \textcolor{blue}{
        \FOR{desired number of pretraining steps}
            \STATE Sample a mini-batch $B = \{(s, a)\}$ from $\mathcal{D}$
            \STATE Update RND predictor weights $\psi$ with gradient descent using
            \begin{equation*}
                \nabla_{\psi} \frac{1}{|B|} \sum_{s \in B} \Bigr[ \lVert f_{\psi}(s, a) - \bar{f}_{\bar{\psi}}(s, a) \rVert_{2} ^ 2  \Bigr]
            \end{equation*}
        \ENDFOR
    }

    \FOR{desired number of training steps}
        \STATE Sample a mini-batch $B = \{(s, a, r, s')\}$ from $\mathcal{D}$
        \STATE Compute target Q-values (shared by all Q-functions):
        \begin{equation*}
            y(r, s') = r + \gamma \Bigr[ \min_{j=1,2} Q_{\bar{\phi}_{i}}(s', a') - \beta \log\pi_{\theta}(a' | s') - \textcolor{blue}{\alpha b(s', a')} \Bigr]
        \end{equation*}
        where $a' \sim \pi_{\theta}(\cdot | s')$ and \textcolor{blue}{$b(s', a')$} is an anti-exploration bonus defined by Eq. (\ref{eq:offline-rnd-bonus}).
        
        \STATE Update each Q-function $Q_{\phi_{i}}$ with gradient descent using
        \begin{equation*}
            \nabla_{\phi_{i}} \frac{1}{|B|} \sum_{(s, a, r, s') \in B} \Bigr[ Q_{\phi_{i}}(s, a) - y(r, s') \Bigr]^2
        \end{equation*}
        \STATE Update policy with gradient ascent using
        \begin{equation*}
            \nabla_{\theta} \frac{1}{|B|} \sum_{s \in B} \Bigr[ \min_{j=1,2} Q_{\phi_{i}}(s, \tilde{a}_{\theta}(s)) - \beta \log \pi(\tilde{a}_{\theta}(s) | s) - \textcolor{blue}{\alpha b(s, \tilde{a}_{\theta}(s))} \Bigr]
        \end{equation*}
        where $\tilde{a}_{\theta}(s)$ is a sample from $\pi(\cdot | s)$ which is differentiable w.r.t. $\theta$ via the reparametrization trick.
        \STATE Update target networks with $\bar{\phi}_{i} \leftarrow (1 - \rho) \bar{\phi_{i}} + \rho \phi_{i}$
    \ENDFOR
\end{algorithmic}
\end{algorithm*}

\begin{algorithm*}[h]
   \caption{Simplified SAC-RND (without a critic) used in experiments for \cref{demonstration} and \cref{exp:cond-pairs}.}
   \label{alg:bc-rnd}
\begin{algorithmic}
    \STATE Initialize policy parameters $\theta$, RND predictor and prior parameters $\{\psi, \psi'\}$, and offline replay buffer $\mathcal{D}$ 
    \FOR{desired number of pretraining steps}
        \STATE Sample a mini-batch $B = \{(s, a)\}$ from $\mathcal{D}$
        \STATE Update RND predictor weights $\psi$ with gradient descent using
        \begin{equation*}
            \nabla_{\psi} \frac{1}{|B|} \sum_{s \in B} \Bigr[ \lVert f_{\psi}(s, a) - \bar{f}_{\bar{\psi}}(s, a) \rVert_{2} ^ 2  \Bigr]
        \end{equation*}
    \ENDFOR

    \FOR{desired number of training steps}
        \STATE Sample a mini-batch $B = \{(s, a, r, s')\}$ from $\mathcal{D}$
        
        \STATE Update policy with gradient descent using
        \begin{equation*}
            \nabla_{\theta} \frac{1}{|B|} \sum_{s \in B} \Bigr[ \beta \log \pi(\tilde{a}_{\theta}(s) | s) + b(s, \tilde{a}_{\theta}(s)) \Bigr]
        \end{equation*}
        where $\tilde{a}_{\theta}(s)$ is a sample from $\pi(\cdot | s)$ which is differentiable w.r.t. $\theta$ via the reparametrization trick and $b(s', a')$ is an anti-exploration bonus defined by Eq. (\ref{eq:offline-rnd-bonus}).
    \ENDFOR
\end{algorithmic}
\end{algorithm*}


\end{document}